%% file: main.tex
\documentclass[english]{article}
\usepackage[T1]{fontenc}
\usepackage[latin9]{inputenc}
\usepackage{xcolor}
\usepackage{babel}
\usepackage{array}
\usepackage{float}
\usepackage{booktabs}
\usepackage{textcomp}
\usepackage{multirow}
\usepackage{amsmath}
\usepackage{amsthm}
\usepackage{amssymb}
\usepackage{graphicx}
\PassOptionsToPackage{normalem}{ulem}
\usepackage{ulem}
\usepackage[unicode=true]
 {hyperref}

\makeatletter

\providecommand{\tabularnewline}{\\}

\PassOptionsToPackage{numbers, compress}{natbib}

\usepackage[preprint]{neurips_2024}

\PassOptionsToPackage{numbers, compress}{natbib}

\usepackage[unicode=true]{hyperref}       
\usepackage{url}            
\usepackage{booktabs}       
\usepackage{amsfonts}       
\usepackage{nicefrac}       
\usepackage{microtype}      
\usepackage{xcolor} 

\usepackage{microtype}
\usepackage{graphicx}
\usepackage{mathtools}

\usepackage{xcolor}
\usepackage{soul}

\definecolor{red}{HTML}{FF0000}
\definecolor{lawngreen}{HTML}{7CFC00}
\definecolor{orange}{HTML}{FFA500}
\definecolor{peru}{HTML}{CD853F}
\definecolor{gray}{HTML}{808080}
\definecolor{lightgray}{RGB}{240,240,240}

\usepackage{algpseudocode}

\AtBeginDocument{%
  \let\cite\citep
}

\@ifundefined{showcaptionsetup}{}{%
 \PassOptionsToPackage{caption=false}{subfig}}
\usepackage{subfig}
\makeatother

\begin{document}
\renewcommand{\contentsname}{Table of Content for Appendix}
\title{Universal Multi-Domain Translation via \\
Diffusion Routers}
\author{\textbf{Duc Kieu$^{1,\ast}$,\enskip{} Kien Do$^{1,\ast}$,\enskip{}
Tuan Hoang$^{1}$,\enskip{} Thao Minh Le$^{2}$,\enskip{} Tung Kieu$^{3}$,
}\\
\textbf{ Dang Nguyen$^{1}$,\quad{} Thin Nguyen$^{1}$}\\
\textbf{$^{1}$}Applied Artificial Intelligence Initiative, Deakin
University, Australia\\
\textbf{$^{2}$}Pennsylvania State University, USA\textbf{\enskip{}}$^{3}$Aalborg
University, Denmark\\
\textbf{$^{1}$}\emph{\{v.kieu, k.do, tuan.h, d.nguyen, thin.nguyen\}@deakin.edu.au
}\\
\emph{$^{2}$mxl6224@psu.edu }\textbf{\qquad{}\qquad{}\qquad{}
}\emph{$^{3}$tungkvt@cs.aau.dk}}

\maketitle
\global\long\def\Expect{\mathbb{E}}%
\global\long\def\Real{\mathbb{R}}%
\global\long\def\Data{\mathcal{D}}%
\global\long\def\Loss{\mathcal{L}}%
\global\long\def\Normal{\mathcal{N}}%
\global\long\def\softmax{\text{softmax}}%
\global\long\def\ELBO{\text{ELBO}}%
\global\long\def\argmin#1{\underset{#1}{\text{argmin}}}%
\global\long\def\argmax#1{\underset{#1}{\text{argmax}}}%
\def\thefootnote{*}\footnotetext{Equal contribution}
\begin{abstract}
\input{abstract.tex}

\addtocontents{toc}{\protect\setcounter{tocdepth}{-1}}
\end{abstract}

\section{Introduction}

\input{intro.tex}

\section{Preliminaries\label{sec:Preliminaries}}

\input{prelim.tex}

\section{Universal Multi-Domain Translation}

\input{problem.tex}

\section{Diffusion Routers}

\input{method.tex}

\section{Experiment\label{sec:Experiment}}

\vspace{-0.2em}\input{exp.tex}\vspace{-0.2em}

\section{Related Work}

\vspace{-0.2em}\input{relate.tex}\vspace{-0.2em}

\section{Conclusion}

\vspace{-0.2em}\input{discuss.tex}

\clearpage{}

\bibliographystyle{plain}
\bibliography{reference}

\addtocontents{toc}{\protect\setcounter{tocdepth}{2}}

\clearpage{}

\appendix
\tableofcontents{}

\clearpage{}

\input{appdx.tex}

\end{document}

%% file: abstract.tex
Multi-domain translation (MDT) aims to learn translations between
multiple domains, yet existing approaches either require fully aligned
tuples or can only handle domain pairs seen in training, limiting
their practicality and excluding many cross-domain mappings. We introduce
universal MDT (UMDT), a generalization of MDT that seeks to translate
between any pair of $K$ domains using only $K-1$ paired datasets
with a central domain. To tackle this problem, we propose Diffusion
Router (DR), a unified diffusion-based framework that models all central$\leftrightarrow$non-central
translations with a single noise predictor conditioned on the source
and target domain labels. DR enables indirect non-central translations
by routing through the central domain. We further introduce a novel
scalable learning strategy with a variational-bound objective and
an efficient Tweedie refinement procedure to support direct non-central
mappings. Through evaluation on three large-scale UMDT benchmarks,
DR achieves state-of-the-art results for both indirect and direct
translations, while lowering sampling cost and unlocking novel tasks
such as sketch$\leftrightarrow$segmentation. These results establish
DR as a scalable and versatile framework for universal translation
across multiple domains.

%% file: intro.tex
Paired domain translation, which aims to learn a mapping between two
domains given aligned samples, underpins a wide range of applications,
including image-to-image translation \cite{isola2017image,park2019semantic},
image captioning \cite{xu2015show,fang2015captions}, and text-to-speech
synthesis \cite{van2016wavenet,shen2018natural}. Despite remarkable
progress in two-domain settings, many real-world problems inherently
involve multiple domains, motivating the study of \textit{Multi-Domain
Translation} (MDT).

Existing MDT approaches usually fall into two paradigms: (i) training
on fully aligned tuples across domains \cite{wu2018multimodal,shi2019variational,bao2023one,le2025one},
or (ii) training on multiple paired datasets with a shared central
domain \cite{Huang2022,zhang2023adding,huang2023composer,koley2024s}.
The former quickly becomes impractical as the number of domains grows
due to the difficulty of collecting large-scale aligned tuples. The
latter scales better but mainly supports translations between the
central domain and each non-central domain, leaving cross non-central
translations unaddressed.

In this paper, we introduce \textit{Universal Multi-Domain Translation}
(UMDT), which combines the ambition of enabling translations between
\emph{any} pairs of $K$ domains with the practicality of only requiring
$K-1$ paired datasets involving a central domain. UMDT captures many
real-world scenarios, such as image$\leftrightarrow$text$\leftrightarrow$audio
translation or multilingual machine translation, where fully aligned
tuples are scarce but pairwise datasets with a pivot domain (e.g.,
text or English) are abundant.

To address UMDT, we propose \textit{Diffusion Router} (DR), a novel
diffusion-based framework that supports arbitrary cross-domain translations
with only a single noise prediction network $\epsilon_{\theta}$.
Inspired by network routers that determine paths using source and
destination IP addresses, DR conditions $\epsilon_{\theta}$ on both
source and target domain labels, guiding the denoising process along
the correct translation path. This design avoids training a separate
model for each mapping, enabling scalability to large numbers of domains.
DR can perform \textit{indirect} translation between non-central domains
via the central domain. To further enable \emph{direct} non-central
translations, we introduce a scalable learning strategy that minimizes
a variational upper bound on the KL divergence between indirect and
parameterized direct mappings. This reduces to aligning two conditional
noise predictions--one conditioned on the source non-central-domain
sample and the other on its paired central-domain sample--resembling
the training objectives of diffusion models. To improve efficiency,
we develop Tweedie refinement, a lightweight sampling procedure that
approximates conditional samples in only a few steps, greatly reducing
computational cost and facilitating scalable training.

Extensive experiments and ablation studies on three newly constructed
large-scale UMDT datasets demonstrate that DR consistently outperforms
state-of-the-art GAN-, flow-, and diffusion-based baselines for multi-domain
translations, either indirect or direct. Moreover, DR naturally generalize
to more complex UMDT topologies such as spanning trees \textcolor{black}{with
multiple central domains}.

In summary, our main contributions are:
\begin{itemize}
\item We formalize Universal Multi-Domain Translation (UMDT), a general
setting that aims to learn translations between \emph{any} pairs of
$K$ domains using only $K-1$ paired datasets. 
\item We propose the Diffusion Router (DR), a unified diffusion-based framework
that models all central$\leftrightarrow$non-central mappings with
a single noise predictor.
\item We develop a scalable learning strategy with a variational-bound objective
and Tweedie refinement to enable direct non-central translations.
\item We construct three new UMDT benchmarks and show that DR achieves state-of-the-art
results for both indirect and direct translations.
\end{itemize}

%% file: prelim.tex
\subsection{Diffusion Models}

Diffusion models \cite{sohl2015deep,song2019generative,ho2020denoising,song2020denoising}
are a generative framework that generates data from noise by reversing
a forward diffusion process. The forward process is typically a Markov
process with the transition kernel $p\left(x_{t}|x_{t-1}\right)$
chosen such that the marginal distribution $p\left(x_{t}|x\right)$
(with $x_{0}\equiv x$) is a Gaussian distribution of the form $\mathcal{N}\left(a_{t}x,\sigma_{t}^{2}\mathrm{I}\right)$.
This enables direct sampling of $x_{t}$ from $x$ as:
\begin{equation}
x_{t}=a_{t}x+\sigma_{t}\epsilon\label{eq:diff_fwd_xt}
\end{equation}
Here, $\epsilon\sim\mathcal{N}\left(0,\mathrm{I}\right)$; $a_{t}$
and $\sigma_{t}$ are predefined time-dependent coefficients satisfying
$1\approx a_{0}>\dots>a_{T}\approx0$ and $0\approx\sigma_{0}^{2}<\dots<\sigma_{T}^{2}\approx1$.
At the final time step, $p\left(x_{T}\mid x\right)\approx p\left(x_{T}\right)=\mathcal{N}\left(0,\mathrm{I}\right)$.

It has been shown that \cite{song2020denoising} the reverse transition
kernel $p_{\theta}\left(x_{t-1}|x_{t}\right)$ can be parameterized
as a Gaussian distribution $\mathcal{N}\left(\mu_{\theta,t,t-1}\left(x_{t}\right),\omega_{t-1|t}^{2}\mathrm{I}\right)$
where the mean is defined as:
\begin{equation}
\mu_{\theta,t,t-1}\left(x_{t}\right)=\frac{a_{t-1}}{a_{t}}x_{t}+\left(\sqrt{\sigma_{t-1}^{2}-\omega_{t-1|t}^{2}}-\frac{\sigma_{t}a_{t-1}}{a_{t}}\right)\epsilon_{\theta}\left(x_{t},t\right),\label{eq:diff_bck_mean}
\end{equation}
and the variance is ${\displaystyle \omega_{t-1|t}^{2}=\eta^{2}\sigma_{t-1}^{2}\left(1-\frac{\sigma_{t-1}^{2}}{\sigma_{t}^{2}}\frac{a_{t}^{2}}{a_{t-1}^{2}}\right)}$
with $\eta\in\left[0,1\right]$ \cite{song2020denoising}.

The noise prediction network $\epsilon_{\theta}$ is trained to predict
the noise $\epsilon$ used to construct $x_{t}$ in the forward process
(see Eq.~\ref{eq:diff_fwd_xt}), using the following loss \cite{ho2020denoising}:

\begin{equation}
\mathcal{L}\left(\theta\right)=\mathbb{E}_{x,t,\epsilon}\left[\left\Vert \epsilon_{\theta}\left(x_{t},t\right)-\epsilon\right\Vert _{2}^{2}\right],\label{eq:diff_loss}
\end{equation}

\noindent where $x\sim p\left(x\right)$, $t\sim\mathcal{U}\left(1,T\right)$\footnote{$\mathcal{U}\left(1,T\right)$ denotes a uniform distribution of time
steps between 1 and $T$.}, and $\epsilon\sim\mathcal{N}\left(0,\mathrm{I}\right)$. This objective
can be interpreted as a variational upper bound on $-\log p\left(x\right)$
\cite{sohl2015deep,ho2020denoising,song2020score}.

While diffusion models were initially proposed to model unconditional
data distributions $p\left(x\right)$, they can be extended to conditional
distributions $p\left(x|y\right)$ \cite{dhariwal2021diffusion,ho2021classifier,rombach2022high},
by incorporating the condition $y$ into the noise predictor, i.e.,
$\epsilon_{\theta}\left(x_{t},t,y\right)$. This conditioning can
be implemented either by concatenating $y$ with $x_{t}$ \cite{saharia2022image},
or through cross attention between $x_{t}$ and $y$ \cite{rombach2022high}.

%% file: problem.tex
\begin{figure*}
\begin{centering}
{\resizebox{\textwidth}{!}{%
\par\end{centering}
\begin{centering}
\begin{tabular}[b]{>{\centering}p{0.4\textwidth}>{\centering}p{0.01\textwidth}>{\centering}p{0.4\textwidth}}
\includegraphics[width=0.42\textwidth]{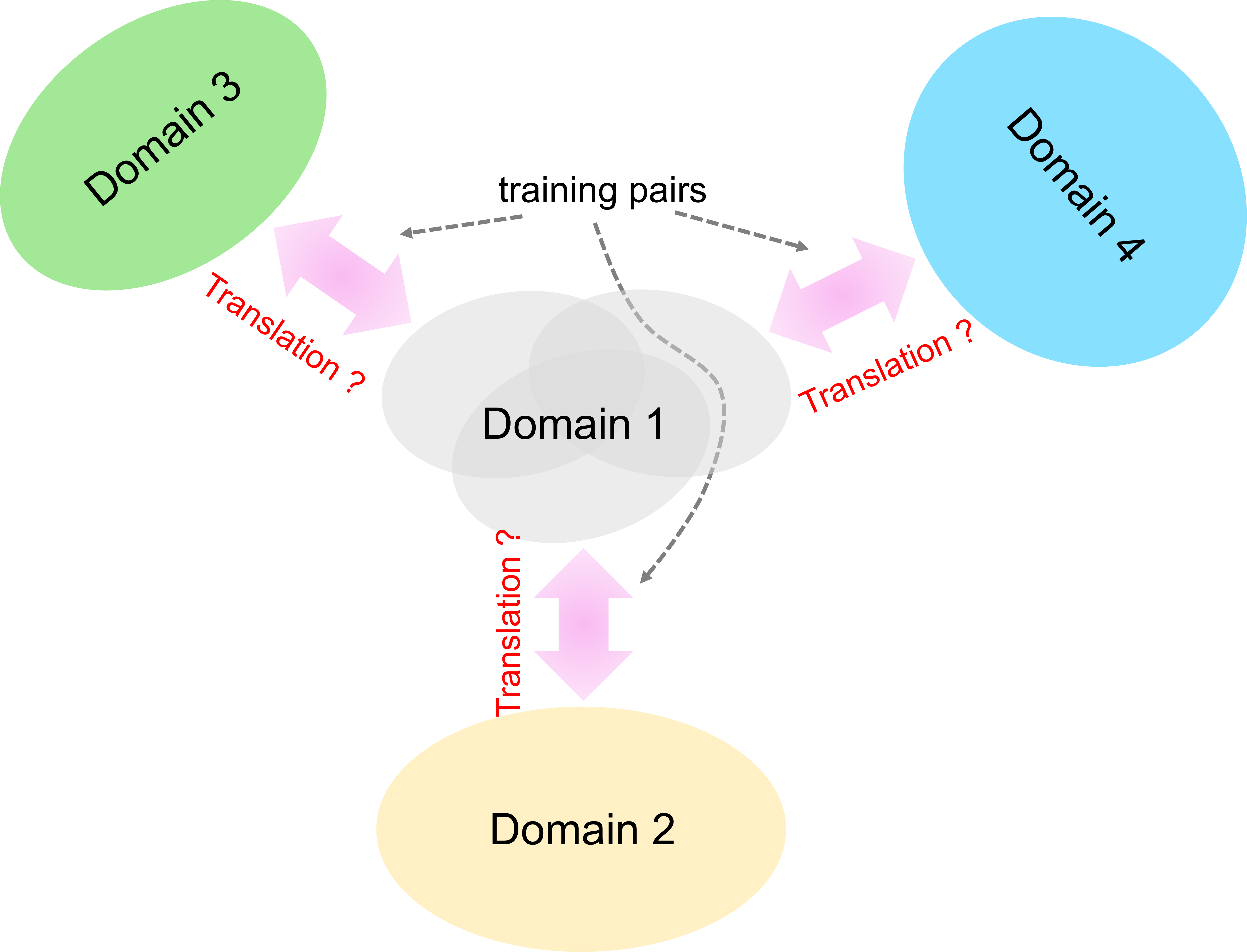} &  & \includegraphics[width=0.42\textwidth]{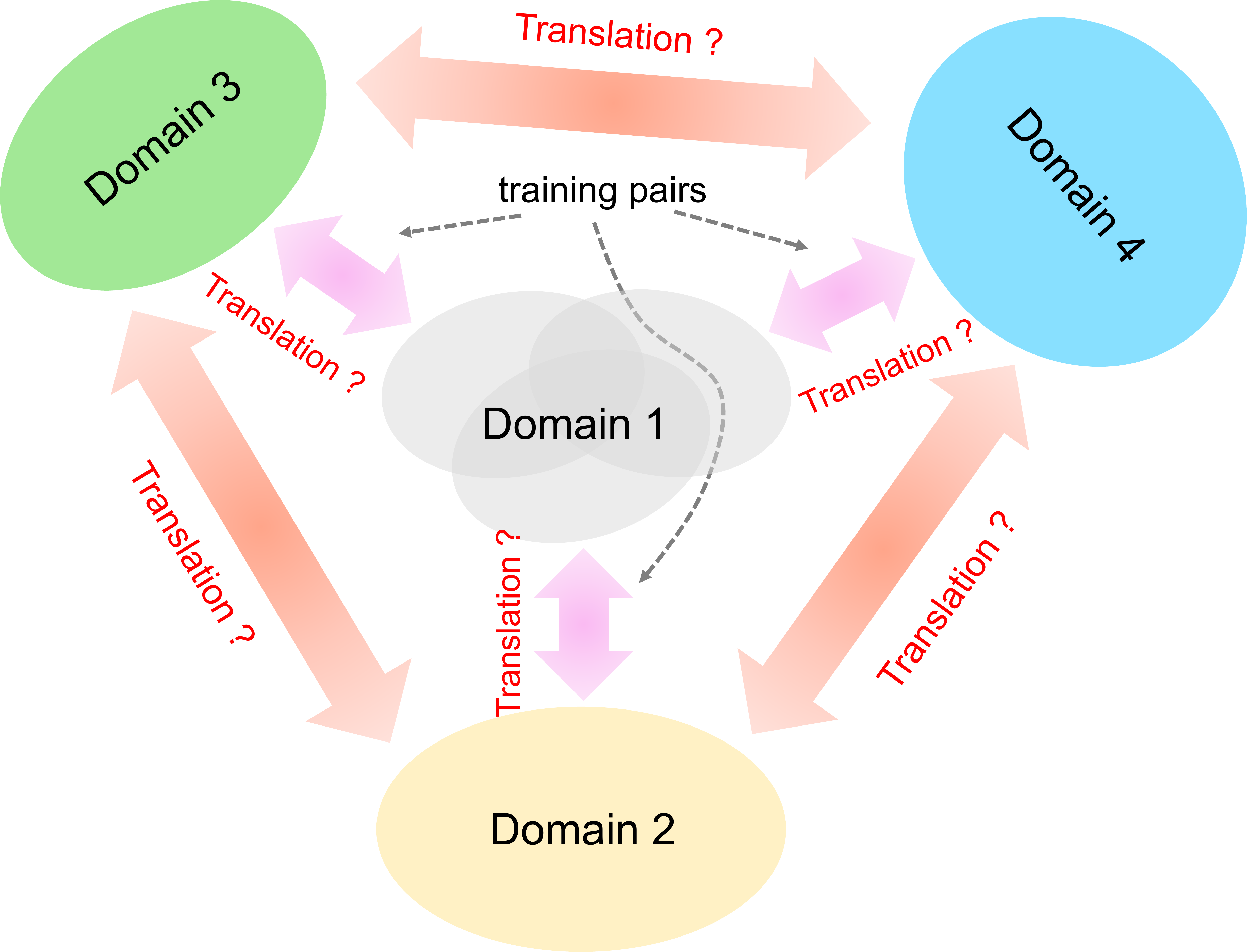}\tabularnewline
{\footnotesize{}(a) Conventional multi-domain translation} &  & {\footnotesize{}(b) Universal multi-domain translation}\tabularnewline
\end{tabular}}}
\par\end{centering}
\caption{Illustration of conventional and universal multi-domain translation\label{fig:Illustration-of-conventional}}
\end{figure*}

We study a more general and challenging extension of conventional
multi-domain translation (MDT) problem (Fig.~\ref{fig:Illustration-of-conventional}a),
which we term \emph{universal multi-domain translation} (UMDT) (Fig.~\ref{fig:Illustration-of-conventional}b).
In this setting, we consider $K$ distinct domains, $X^{1}$, $X^{2}$,
..., $X^{K}$, with training data consisting of $K-1$ paired datasets
between each domain $X^{k}$ and a \emph{shared central domain} $X^{c}$
(where $1\leq c\leq K$): $\mathcal{D}_{k,c}=\left\{ \left(x^{k},x^{c}\right)\right\} _{n=1}^{N_{k}}$,
for all $k\neq c$. The goal is to learn a model that can translate
between \emph{any} pair of domains $\left(X^{i},X^{j}\right)$ for
\emph{all} $i\neq j$. Since paired data between non-central domains
are unavailable, we assume that samples from the central domain $X^{c}$
share overlapping information across the paired datasets. This is
a mild assumption and is typically satisfied in practice.

UMDT is highly practical. For example, in multi-modal translation
across images, text, and audio, it is often difficult to obtain large-scale
datasets with fully aligned triplets (image, text, audio). However,
paired datasets such as image-text (e.g., image captions) and text-audio
(e.g., audiobooks) are more common. In this case, text naturally serve
as the central domain, with image and audio as non-central domains.
Image$\leftrightarrow$audio translation can then be achieved indirectly
via text, even without direct image-audio training pairs. Importantly,
text samples across datasets \emph{need not match exactly}; loose
overlaps--such as shared vocabulary or semantics--are sufficient~for~image$\leftrightarrow$audio~translation.

More generally, UMDT can be extended to cases where the $K-1$ paired
datasets form a \emph{spanning tree} over the $K$ domains. Since
spanning trees can take arbitrary structures, in this work we restrict
our attention to the star-shaped configuration. Nonetheless, the method
proposed in the following section naturally generalizes to the broader
spanning-tree setting.

%% file: method.tex
\subsection{Indirect translation between non-central domains via the central
one\label{subsec:Indirect-translation}}

From a probabilistic perspective, bidirectional translation between
two non-central domains $X^{i}$ and $X^{j}$ can be viewed as sampling
from the conditional distributions $p\left(x^{i}|x^{j}\right)$ and
$p\left(x^{j}|x^{i}\right)$. These can be expressed via the central
domain $X^{c}$ as:
\begin{equation}
p\left(x^{j}|x^{i}\right)=\int p\left(x^{j}|x^{c}\right)p\left(x^{c}|x^{i}\right)dx^{c},\ \ \ p\left(x^{i}|x^{j}\right)=\int p\left(x^{i}|x^{c}\right)p\left(x^{c}|x^{j}\right)dx^{c}\label{eq:cond_distr}
\end{equation}
Here, we assume $X^{i}\perp X^{j}|X^{c}$ for all $i,j\neq c$ which
leads to $p\left(x^{i}|x^{c}\right)=p\left(x^{i}|x^{c},x^{j}\right)$
and $p\left(x^{j}|x^{c}\right)=p\left(x^{j}|x^{c},x^{i}\right)$.
In other words, once the central domain $X^{c}$ is known, the non-central
domains become conditionally independent of each other.

This formulation extends the transitivity property of equivalence
relations into a probabilistic framework. It implies that if we can
learn bidirectional mappings between the central domain $X^{c}$ and
each non-central domain $X^{k}$ ($k\neq c$) that capture the couplings
$p\left(x^{k}|x^{c}\right)$ and $p\left(x^{c}|x^{k}\right)$ from
the training data (i.e., solving the conventional MDT problem), then
we can perform indirect translation between any pair $\left(X^{i},X^{j}\right)$
through $X^{c}$, even without direct supervision between $X^{i}$
and $X^{j}$.

Notably, $p\left(x^{k}|x^{c}\right)$ and $p\left(x^{c}|x^{k}\right)$
can be effectively modeled using conditional diffusion models as discussed
in Section~\ref{sec:Preliminaries} or diffusion bridges \cite{LiuVHTNA23,LiX0L23,zhou2024denoising,Kieu2025}
(with details in Appdx.~\ref{subsec:prelim-diffusion-bridges}).
However, naively constructing a separate model for each distribution
would require up to $2(K-1)$ models to cover all translations between
$X^{c}$ and the non-central domains $\left\{ X^{k}|k\neq c\right\} $,
which becomes impractical as the number of domains grows. 

To address this scalability challenge, we propose a unified framework
called \emph{Diffusion Router} (DR) that learns all \emph{bidirectional}
mappings between the central and non-central domains using a \emph{single}
network, thereby avoiding redundancy and enabling efficient multi-domain
translation.

Inspired by network routers, which rely on source and destination
IP addresses to determine the routing path, our framework incorporates
the labels (or indices) of the source and target domains into the
network $\epsilon_{\theta}$. This design allows $\epsilon_{\theta}$
to infer the correct translation path for a given noisy input $x_{t}$.
Specifically, $\epsilon_{\theta}$ takes the form $\epsilon_{\theta}\left(x_{t}^{\text{tgt}},t,x^{\text{src}},\text{tgt},\text{src}\right)$
where $\text{src}$ and $\text{tgt}$ denote the source and target
domain labels. We train DR with the following objective function:
\begin{align}
\mathcal{L}_{\text{paired}}\left(\theta\right) & =\mathbb{E}_{\left(x^{k},x^{c}\right)\sim\mathcal{D}_{k,c},t,\epsilon,\zeta}\bigg[\zeta\left\Vert \epsilon_{\theta}\left(x_{t}^{k},t,x^{c},k,c\right)-\epsilon\right\Vert _{2}^{2}+\left(1-\zeta\right)\left\Vert \epsilon_{\theta}\left(x_{t}^{c},t,x^{k},c,k\right)-\epsilon\right\Vert _{2}^{2}\bigg],\label{eq:paired_loss}
\end{align}
where $t\sim\mathcal{U}\left(1,T\right)$, $\epsilon\sim\Normal\left(0,\mathrm{I}\right)$,
and $\zeta\sim\mathcal{B}\left(0.5\right)$\footnote{$\mathcal{B}\left(0.5\right)$ denotes a Bernoulli distribution with
the probability of getting 1 equal 0.5.}. Next, $x_{t}^{k}=a_{t}x^{k}+\sigma_{t}\epsilon$ for the \emph{standard}
(or diffusion-based) DR variant and $x_{t}^{k}=\alpha_{t}x^{k}+\beta_{t}x^{c}+\sigma_{t}\epsilon$
for the \emph{bridge-based} DR variant.

Once trained, DR can translate between two non-central domains $X^{i}$
and $X^{j}$ indirectly via $X^{c}$, using a \emph{two-stage} process:
1) generate central-domain samples $x^{c}$ conditioned on a source
sample $x^{i}$ (or $x^{j}$), then 2) generate target samples $x^{j}$
(or $x^{i}$) conditioned on the intermediate samples $x^{c}$.

\subsection{Direct translation between non-central domains\label{subsec:Direct-translation}}

Indirect translation requires generating intermediate samples $x^{c}$
of the central domain, which is computationally expensive and sensitive
to sample quality. To overcome these drawbacks, we propose a novel
approach that enables \emph{direct} translation between non-central
domains $X^{i}$ and $X^{j}$ by explicitly modeling $p_{\theta}\left(x^{j}|x^{i}\right)$
and $p_{\theta}\left(x^{i}|x^{j}\right)$. Our method can either finetune
a DR pretrained with Eq.~\ref{eq:paired_loss} or train a new DR
from scratch for direct \emph{}cross-domain translation.

To learn $p_{\theta}\left(x^{j}|x^{i}\right)$ (or similarly $p_{\theta}\left(x^{i}|x^{j}\right)$),
we minimize the following KL divergence:
\begin{align}
 & \mathbb{E}_{p\left(x^{i}\right)}\left[D_{KL}\left[p\left(x^{j}|x^{i}\right)\Vert p_{\theta}\left(x^{j}|x^{i}\right)\right]\right]\nonumber \\
=\  & \Expect_{\left(x^{i},x^{c}\right)\sim\mathcal{D}_{i,c}}\Expect_{p\left(x^{j}|x^{c}\right)}\left[\log\left(\Expect_{p\left({x'}^{c}|x^{i}\right)}\left[p\left(x^{j}|{x'}^{c}\right)\right]\right)-\log p_{\theta}\left(x^{j}|x^{i}\right)\right],\label{eq:orig_KL_div}
\end{align}
where ${x'}^{c}$ denotes samples of $X^{c}$ that are distinct from
$x^{c}$. The detailed derivation of Eq.~\ref{eq:orig_KL_div} is
provided in Appdx.~\ref{subsec:Mathematical-Derivation-of}. The
main bottleneck here is the term $\log\left(\Expect_{p\left({x'}^{c}|x^{i}\right)}\left[p\left(x^{j}|{x'}^{c}\right)\right]\right)$.
First, sampling from $p\left({x'}^{c}|x^{i}\right)$ typically requires
hundreds to thousands of denoising steps if using a pretrained DR.
Second, even if we obtain samples ${x'}^{c}$ from $p\left({x'}^{c}|x^{i}\right)$,
evaluating $p\left(x^{j}|{x'}^{c}\right)$ remains intractable due
to the lack of a closed-form expression.

To overcome this, we approximate $\Expect_{p\left({x'}^{c}|x^{i}\right)}\left[p\left(x^{j}|{x'}^{c}\right)\right]$
by $p\left(x^{j}|x^{c}\right)$ where $x^{c}$ comes from the pair
$\left(x^{i},x^{c}\right)\sim\mathcal{D}_{i,c}$. This leads to the
following tractable training objective:
\begin{align}
 & \Expect_{\left(x^{i},x^{c}\right)\sim\mathcal{D}_{i,c}}\Expect_{p\left(x^{j}|x^{c}\right)}\left[\log\left(\Expect_{p\left({x'}^{c}|x^{i}\right)}\left[p\left(x^{j}|{x'}^{c}\right)\right]\right)-\log p_{\theta}\left(x^{j}|x^{i}\right)\right]\nonumber \\
\approx\  & \Expect_{\left(x^{i},x^{c}\right)\sim\mathcal{D}_{i,c}}\Expect_{p\left(x^{j}|x^{c}\right)}\left[\log p\left(x^{j}|x^{c}\right)-\log p_{\theta}\left(x^{j}|x^{i}\right)\right]\label{eq:log_MC_estimate_1}\\
=\  & \Expect_{\left(x^{i},x^{c}\right)\sim\mathcal{D}_{i,c}}\left[D_{KL}\left(p\left(x^{j}|x^{c}\right)\Vert p_{\theta}\left(x^{j}|x^{i}\right)\right)\right]\label{eq:log_MC_estimate_2}
\end{align}
Although this approximation introduces bias due to the logarithm operating
on a (single-sample) Monte Carlo estimate, we empirically observe
that its impact on learning is manageable, particularly when the conditional
distribution $p\left(x'^{c}|x^{i}\right)$ is sharply peaked at its
mode.

The objective in Eq.~\ref{eq:log_MC_estimate_2} suggests that we
can learn a direct mapping from $X^{i}$ to $X^{j}$ by ensuring that
if $x^{i}$ and $x^{c}$ are semantically aligned (i.e., appear as
a pair in $\mathcal{D}_{i,c}$), then the conditional distribution
over the target domain $X^{j}$ given $x^{i}$ should closely match
that given $x^{c}$. Note that $p\left(x^{j}|x^{c}\right)$ can be
viewed as the path distribution of the stochastic process $X^{c}\rightarrow X^{j}$,
modeled using a \emph{pretrained} DR introduced in Section~\ref{subsec:Indirect-translation}.
Likewise, $p_{\theta}\left(x^{j}|x^{i}\right)$ represents the path
distribution for the process $X^{i}\rightarrow X^{j}$, which we also
model using the same DR. This is implemented by providing $i$, $j$
as inputs to the noise prediction network $\epsilon_{\theta}$ of
the DR. As a result, learning $p_{\theta}\left(x^{j}|x^{i}\right)$
corresponds to finetuning this pretrained DR to support direct mapping
from $X^{i}$ to $X^{j}$.

Let $p_{\text{ref}}\left(x^{j}|x^{c}\right)$ denote the parameterization
of $p\left(x^{j}|x^{c}\right)$ given by the pretrained DR. Instead
of minimizing the KL divergence between $p_{\text{ref}}\left(x^{j}|x^{c}\right)$
and $p_{\theta}\left(x^{j}|x^{i}\right)$ directly, we minimize the
sum of KL divergences between their respective transition kernels.
This sum acts as a variational upper bound on the original KL objective,
as shown in Appdx.~\ref{subsec:Variational-Upper-Bound}:
\begin{align}
 & \mathbb{E}_{\left(x^{i},x^{c}\right)\sim\mathcal{D}_{i,c}}\left[D_{KL}\left(p_{\text{ref}}\left(x^{j}|x^{c}\right)\Vert p_{\theta}\left(x^{j}|x^{i}\right)\right)\right]\nonumber \\
\le\  & \mathbb{E}_{\left(x^{i},x^{c}\right)\sim\mathcal{D}_{i,c}}\left[\sum_{t=1}^{T}\mathbb{E}_{p_{\text{ref}}\left(x_{t}^{j}\mid x^{c}\right)}\left[D_{KL}\left(p_{\text{ref}}\left(x_{t-1}^{j}|x_{t}^{j},x^{c}\right)\Vert p_{\theta}\left(x_{t-1}^{j}|x_{t}^{j},x^{i}\right)\right)\right]\right]+\text{const}\label{eq:VB_path_KL_div}
\end{align}

Ideally, the sampling distribution in Eq.~\ref{eq:VB_path_KL_div}
should be $p_{\text{ref}}\left(x_{t}^{j}|x^{i}\right)$ so that after
training, $p_{\theta}\left(x_{t-1}^{j}|x_{t}^{j},x^{i}\right)$ can
accurately model the transition dynamics from $X^{i}$ to $X^{j}$.
However, the actual sampling distribution is $p_{\text{ref}}\left(x_{t}^{j}|x^{c}\right)$.
For bridge-based DR, the stochastic processes $X^{i}\to X^{j}$ and
$X^{c}\to X^{j}$ starts from different initial states $x^{i}$ and
$x^{c}$, respectively, making their path distributions, $p_{\text{ref}}\left(x_{t}^{j}|x^{i}\right)$
and $p_{\text{ref}}\left(x_{t}^{j}|x^{c}\right)$ inherently different.
Therefore, bridge-based DR are ill-suited for finetuning with the
objective in Eq.~\ref{eq:VB_path_KL_div}. By contrast, standard
DR provide a more viable solution. Since both $X^{i}\rightarrow X^{j}$
and $X^{c}\rightarrow X^{j}$ originate from the same Gaussian prior,
they share a common stochastic path, allowing $p_{\text{ref}}\left(x_{t}^{j}|x^{c}\right)$
to serve as a proxy for $p_{\text{ref}}\left(x_{t}^{j}|x^{i}\right)$.
For this reason, we restrict our focus to finetuning standard DR.

By applying the standard reparameterization trick for diffusion models
\cite{ho2020denoising}, we reformulate Eq.~\ref{eq:VB_path_KL_div}
as a noise prediction loss:
\begin{equation}
\mathcal{L}_{\text{unpaired}}\left(\theta\right)=\mathbb{E}_{\left(x^{i},x^{c}\right)\sim\mathcal{D}_{i,c},x_{t}^{j}\sim p_{\text{ref}}\left(x_{t}^{j}|x^{c}\right),t,\epsilon}\left[\left\Vert \epsilon_{\theta}\left(x_{t}^{j},t,x^{i},j,i\right)-\epsilon_{\text{ref}}\left(x_{t}^{j},t,x^{c},j,c\right)\right\Vert _{2}^{2}\right],\label{eq:unpaired_loss}
\end{equation}
where $\epsilon_{\text{ref}}$ is the frozen noise prediction network
of the pretrained DR.

To prevent catastrophic forgetting of previously learned mappings
between $X^{c}$ and $X^{k}$, we combine Eq.~\ref{eq:unpaired_loss}
with $\mathcal{L}_{\text{paired}}$ (see Eq.~\ref{eq:paired_loss})
resulting in the final loss:
\begin{align}
\Loss_{\text{final}}\left(\theta\right) & =\lambda_{1}\mathcal{L}_{\text{unpaired}}\left(\theta\right)+\lambda_{2}\mathcal{L}_{\text{paired}}\left(\theta\right)\label{eq:final_loss}
\end{align}
Here, the coefficients $\lambda_{1},\lambda_{2}\geq0$ balance the
trade-off between learning new translations $X^{i}\rightarrow X^{j}$
and preserving existing mappings $X^{k}\leftrightarrow X^{c}$. 

Notably, $\Loss_{\text{final}}\left(\theta\right)$ is highly flexible.
It can train DR \emph{from scratch} by treating $\epsilon_{\text{ref}}$
as an online network with frozen parameters rather than a pretrained
model (Appdx.~\ref{subsec:Training-from-scratch}). Moreover, it
supports in the case where paired domains form a spanning tree with
multiple central domains (see Section~\ref{sec:Experiment}).

\subsubsection{Sampling from conditional distributions with Tweedie refinement\label{subsec:Sampling-from-conditional}}

\begin{figure}
\begin{centering}
\includegraphics[width=1\textwidth]{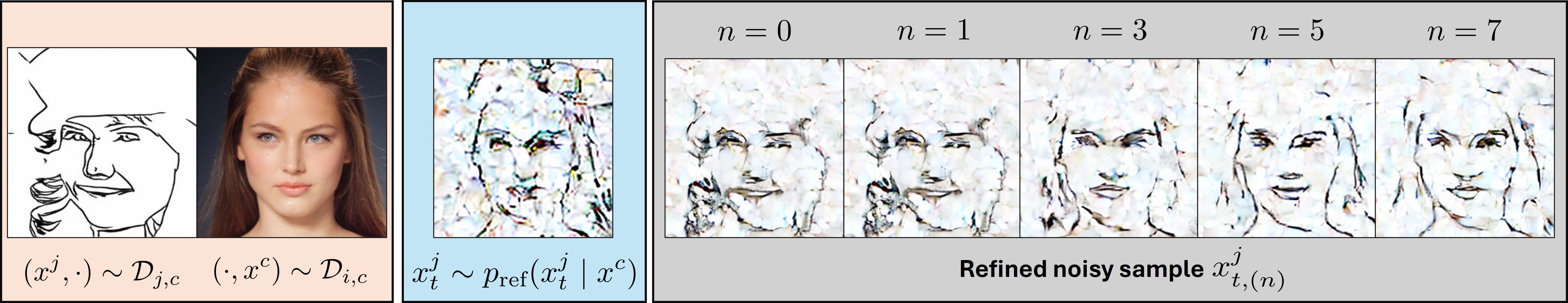}
\par\end{centering}
\caption{Tweedie refinement with $n\in\left\{ 0,1,3,5,7\right\} $ on Faces-UMDT-Latent.
\textbf{Left}: A conditional sample $x^{c}$ and a random target-domain
sample $x^{j}$. \textbf{Middle}: A ground-truth noisy target-domain
sample $x_{t}^{j}$ aligned with $x^{c}$ (\emph{not} available during
training). \textbf{Right}: Tweedie refinement progressively transforms
$x_{t}^{j}\sim p\left(x_{t}^{j}\right)$ into $x_{t}^{j}\sim p\left(x_{t}^{j}|x^{c}\right)$
as $n$ increases.\label{fig:Illustration_tweedie_refine}}
\end{figure}

The main challenge in Eq.~\ref{eq:unpaired_loss} is sampling $x_{t}^{j}$
from $p_{\text{ref}}\left(x_{t}^{j}|x^{c}\right)$. A straightforward
approach is to perform backward denoising from time $T$ to $t$ using
the pretrained DR, but this is computationally expensive and does
not scale well. To address this, we propose a novel sampling method:
\begin{equation}
x_{t,(n+1)}^{j}=x_{t,(n)}^{j}+\sigma_{t}\left(\epsilon-\epsilon_{\theta}\left(x_{t,(n)}^{j},t,x^{c},j,c\right)\right)\label{eq:Tweedie_refinement}
\end{equation}
where $\epsilon\sim\mathcal{N}\left(0,\mathrm{I}\right)$ and $x_{t,(n)}^{j}$
denotes the refined sample after $n$ steps, initialized with $x_{t,(0)}^{j}\sim p_{\text{ref}}\left(x_{t}^{j}\right)$.
A sample $x_{t}^{j}\sim p_{\text{ref}}\left(x_{t}^{j}\right)$ can
be obtained by first drawing $\left(x^{j},\cdot\right)$ from $\mathcal{D}_{j,c}$
and then applying the forward diffusion process $x_{t}^{j}=a_{t}x^{j}+\sigma_{t}\epsilon$. 

We refer to this procedure as \emph{Tweedie refinement} due to its
connection with Tweedie's formula \cite{efron2011tweedie}. Empirically,
we find that Tweedie refinement can approximate samples from $p_{\text{ref}}\left(x_{t}^{j}|x^{c}\right)$
with only a few refinement steps (see Fig.~\ref{fig:Illustration_tweedie_refine}).
Compared with existing refinement techniques \cite{song2020score,Yu2023},
our approach (1) introduces a distinct formulation, (2) converts unconditional
samples into conditional ones rather than projecting off-distribution
samples back onto a marginal distribution, and (3) is applied during
training rather than inference.

%% file: exp.tex
\subsection{Experimental Setup}

\begin{table*}
\begin{centering}
{\resizebox{\textwidth}{!}{%
\par\end{centering}
\begin{centering}
\begin{tabular}{ccccccc}
\toprule 
\multirow{3}{*}{Method} & \multicolumn{6}{c}{FID$\downarrow$}\tabularnewline
\cmidrule{2-7} \cmidrule{3-7} \cmidrule{4-7} \cmidrule{5-7} \cmidrule{6-7} \cmidrule{7-7} 
 & \multicolumn{3}{c}{Shoes-UMDT} & \multicolumn{3}{c}{Faces-UMDT-Latent}\tabularnewline
\cmidrule{2-7} \cmidrule{3-7} \cmidrule{4-7} \cmidrule{5-7} \cmidrule{6-7} \cmidrule{7-7} 
 & Edge$\leftrightarrow$Shoe & Gray.$\leftrightarrow$Shoe & \textcolor{brown}{Edge$\leftrightarrow$Gray.} & Ske.$\leftrightarrow$Face & Seg.$\leftrightarrow$Face & \textcolor{brown}{Ske.$\leftrightarrow$Seg.}\tabularnewline
\midrule
\midrule 
StarGAN & 9.92/20.18 & 19.73/42.61 & \textcolor{brown}{18.64/27.41} & - & - & \textcolor{brown}{-}\tabularnewline
\midrule 
Rectified Flow & 2.88/30.92 & 3.75/43.38 & \textcolor{brown}{20.14/18.83} & 20.22/97.76 & 10.85/81.44 & \textcolor{brown}{50.82/17.31}\tabularnewline
\midrule 
UniDiffuser & 2.98/11.94 & 2.72/4.40 & \textcolor{brown}{4.81/12.26} & 13.13/55.46 & 11.02/46.04 & \textcolor{brown}{36.13/12.52}\tabularnewline
\midrule 
iDR & \textbf{1.66}/\textbf{5.15} & \textbf{0.53}/\textbf{1.60} & \textbf{\textcolor{brown}{1.85}}\textcolor{brown}{/}\textbf{\textcolor{brown}{5.48}} & \textbf{9.07}/\textbf{23.88} & \uline{6.12}/\textbf{19.12} & \textbf{\textcolor{brown}{15.37}}\textcolor{brown}{/}\textcolor{brown}{\uline{6.15}}\tabularnewline
\midrule 
dDR & \uline{2.01}/\uline{5.76} & \uline{0.57}/\uline{1.69} & \textcolor{brown}{\uline{2.74}}\textcolor{brown}{/}\textcolor{brown}{\uline{6.51}} & \uline{9.62}/\uline{27.09} & \textbf{3.43}/\uline{21.26} & \textcolor{brown}{\uline{19.42}}\textcolor{brown}{/}\textbf{\textcolor{brown}{5.52}}\tabularnewline
\bottomrule
\end{tabular}}}
\par\end{centering}
\caption{FID scores on Shoes-UMDT and Faces-UMDT-Latent. Translations without
paired data are marked in \textcolor{brown}{brown}. The best results
are shown in \textbf{bold}, and the second-best are \uline{underlined}.\label{tab:result_Shoes-UMDT}}
\end{table*}

\subsubsection{Datasets}

Since the proposed UMDT problem is novel, no datasets currently exist
for it. To address this, we create three benchmark datasets for evaluating
our method, namely \textbf{Shoes-UMDT}, \textbf{Faces-UMDT}, and \textbf{COCO-UMDT}.
Detailed descriptions of these datasets are provided below.

\paragraph{Shoes-UMDT}

This dataset is adapted from the Edges2Shoes dataset \cite{isola2017image}
used for paired image-to-image translation. From the original 50K
(shoe, edge) pairs, we randomly sample two \emph{disjoint} subsets
of 20K pairs each. One subset remains unchanged, while in the other,
the \textquotedblleft edge\textquotedblright{} image in each pair
is replaced with a ``grayscale'' version of the corresponding \textquotedblleft shoe\textquotedblright{}
image, rotated by 20 degrees and scaled by 20\% smaller . This yields
two disjoint sets: 20K (shoe, edge) pairs and 20K (shoe, grayscale)
pairs. In this setup, ``shoe'' is the central domain, while ``edge''
and ``grayscale'' are non-central domains. The remaining 10K (shoe,
edge) pairs are used to generate grayscale images following the same
procedure as in the training data, producing a test set of 10K (shoe,
edge, grayscale) triplets. All images are resized to 64$\times$64$\times$3.

\paragraph{Faces-UMDT}

We build Faces-UMDT by combining CelebA-Mask-HQ \cite{CelebAMask-HQ}
(30K (face, segment) pairs) and FFHQ \cite{karras2019style} (70K
face images). From FFHQ, we randomly select 30K face images and use
Sobel filter followed by sketch simplifier \cite{simo2018mastering}
to generate corresponding sketches, producing 30K (face, sketch) pairs.
These are merged with 25K randomly selected (face, segment) pairs
from CelebA-Mask-HQ to form the training set in which ``face'' is
the central domain and ``segment'', ``sketch'' are non-central
domains. For testing, we generate a sketch for each face image in
the remaining 5K CelebA-Mask-HQ pairs, resulting in 5K (face, segment,
sketch) triplets.

This setup reflects real-world scenarios where the two disjoint subsets
of face images associated with CelebA-Mask-HQ and FFHQ follow distinct
distributions, making sketch$\leftrightarrow$segment translation
via the face domain more challenging. For this dataset, we consider
two settings: 1) resizing input images to 128$\times$128$\times$3
and translating in the pixel space, and 2) encoding 256$\times$256$\times$3
images using a VAE encoder \cite{rombach2022high} and translating
in the latent space of shape 32$\times$32$\times$4. These settings
result in two versions which are Faces-UMDT-Pixel and Face-UMDT-Latent,
respectively.

\begin{figure*}
\begin{centering}
\vspace{-0.4em}\subfloat[Shoes-UMDT]{\begin{centering}
\includegraphics[width=0.48\textwidth]{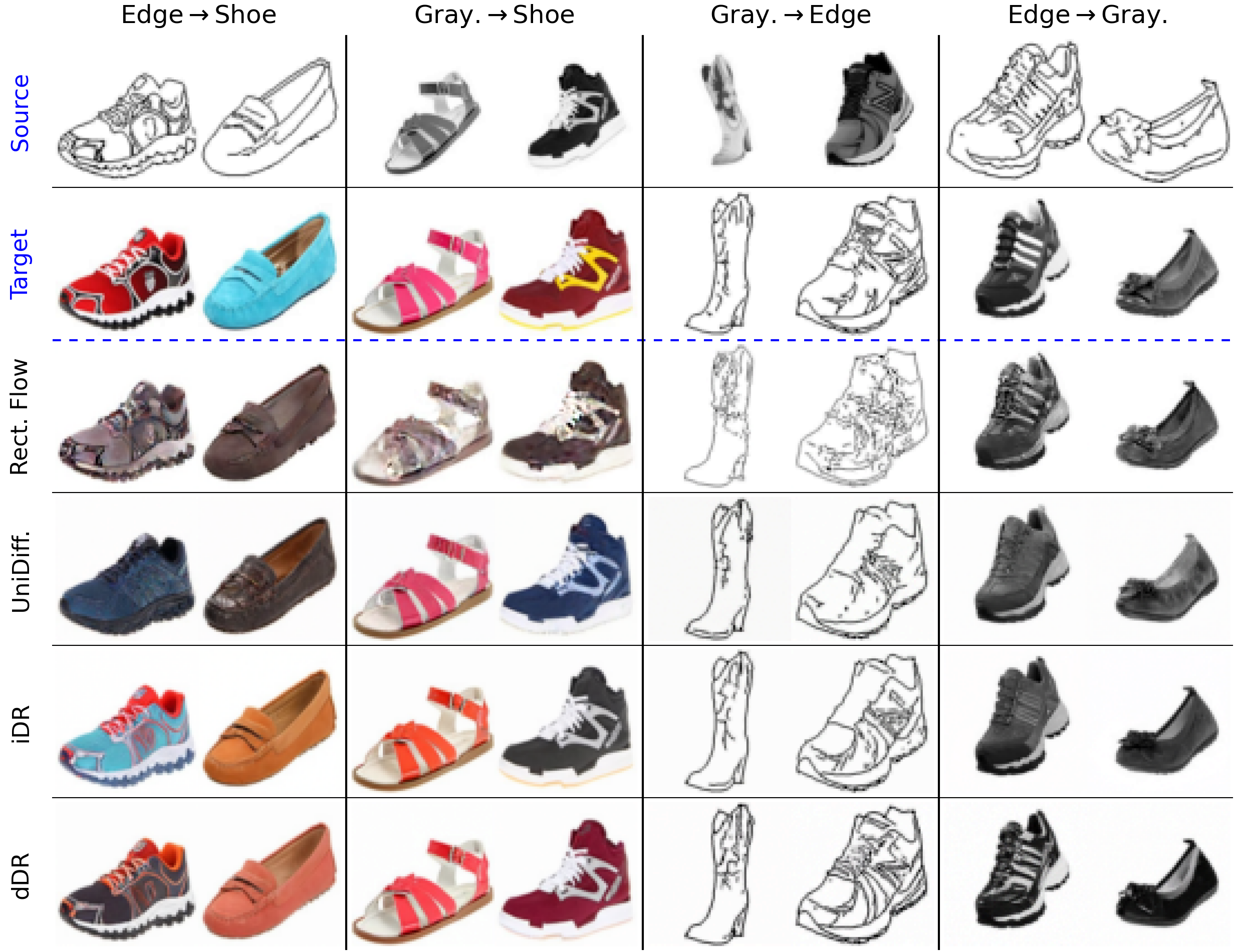}
\par\end{centering}
}\subfloat[Faces-UMDT-Latent]{\begin{centering}
\includegraphics[width=0.48\textwidth]{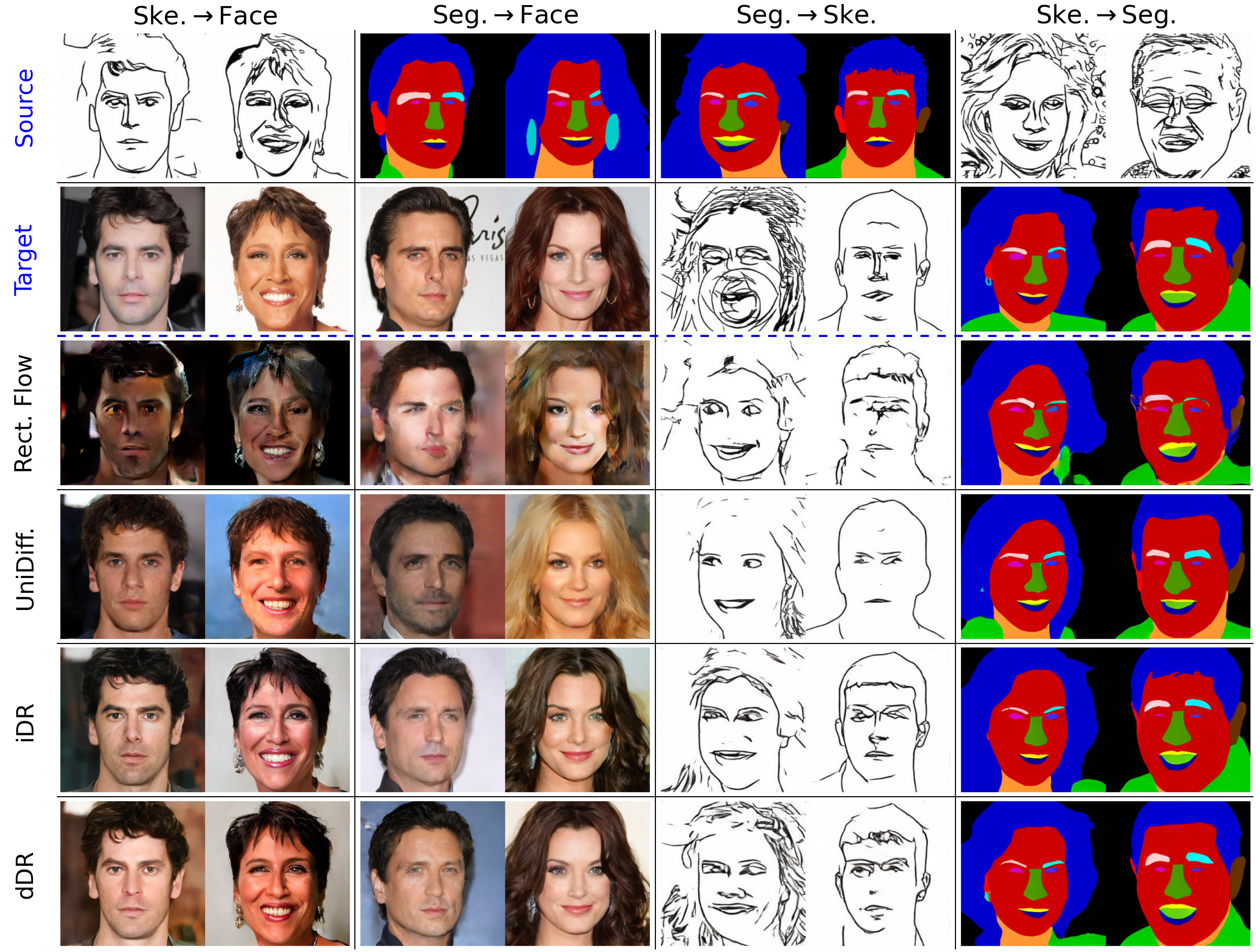}
\par\end{centering}
}
\par\end{centering}
\centering{}\caption{Qualitative results on Shoes-UMDT and Faces-UMDT-Latent.\label{fig:Qualitative-results}}
\vspace{-0.4em}
\end{figure*}

\paragraph{COCO-UMDT-Star and COCO-UMDT-Chain}

COCO-Stuff \cite{Caesar2018} is a large and diverse image segmentation
dataset with 118K (color, segment) pairs for training and 5K pairs
for evaluation, covering 80 \textquotedblleft thing\textquotedblright{}
classes, 91 \textquotedblleft stuff\textquotedblright{} classes, and
one \textquotedblleft unlabeled\textquotedblright{} class. Following
\cite{Mou2024}, we generate additional domains by applying the Pixel
Difference Network \cite{su2021pixel} to extract sketches and MiDaS
\cite{ranftl2020towards} to produce depth maps from all color images.
For COCO-UMDT-Star, we construct three training subsets of 70K color
images each, paired with segmentation maps, sketches, and depth maps,
respectively, yielding (color, segment), (color, sketch), and (color,
depth) training pairs. For COCO-UMDT-Chain, we form three paired subsets:
(segment, color), (color, sketch), and (sketch, depth), each with
70K pairs. Both COCO-UMDT-Star and COCO-UMDT-Chain share a common
test set of 5K (color, segment, sketch, depth) quadruplets derived
from the original evaluation split. All images are resized to 256$\times$256$\times$3
and mapped into latent space of size 32$\times$32$\times$4 using
VAE encoder from \cite{rombach2022high}.

\subsubsection{Baselines and metrics}

We consider two versions of DR. The first is trained with the loss
$\mathcal{L}_{\text{paired}}\left(\theta\right)$ in Eq.~\ref{eq:paired_loss},
which can only perform indirect translations between non-central domains.
The second is finetuned from the first using the loss $\mathcal{L}_{\text{final}}\left(\theta\right)$
in Eq.~\ref{eq:final_loss}, allowing direct cross-domain translation.
We refer to these two versions as iDR and dDR, respectively. We also
train version from scratch using $\mathcal{L}_{\text{final}}\left(\theta\right)$
and compare with the finetuned version in Appdx.~\ref{subsec:Training-from-scratch}.

As with datasets, methods for addressing the UMDT problem remain largely
unexplored. To establish baselines, we adapt several approaches originally
designed for the conventional MDT setting. These include StarGAN \cite{choi2018stargan},
UniDiffuser \cite{bao2023one}, and Rectified Flow \cite{liu2022flow}
as representatives for GAN-based, diffusion-based, and flow-based
methods. Further implementation details for both our method and the
baselines are provided in Appdx.~\ref{subsec:Implementation-Detail}.

Following previous works on MDT, we use FID \cite{heusel2017gans}
and LPIPS \cite{zhang2018perceptual} to measure distributional fidelity
and perceptual similarity, respectively. In all tables, each entry
for translations A$\leftrightarrow$B is reported as X/Y, where X
is the metric for A$\leftarrow$B and Y is the metric for A$\to$B.

\begin{table*}
\begin{centering}
\vspace{-0.3em}{\resizebox{\textwidth}{!}{%
\par\end{centering}
\begin{centering}
\begin{tabular}{ccccccc}
\toprule 
\multirow{2}{*}{Method} & \multicolumn{6}{c}{FID$\downarrow$}\tabularnewline
\cmidrule{2-7} \cmidrule{3-7} \cmidrule{4-7} \cmidrule{5-7} \cmidrule{6-7} \cmidrule{7-7} 
 & Ske.$\leftrightarrow$Color & Seg.$\leftrightarrow$Color & Depth$\leftrightarrow$Color & \textcolor{brown}{Ske.$\leftrightarrow$Seg.} & \textcolor{brown}{Ske.$\leftrightarrow$Depth} & \textcolor{brown}{Seg.$\leftrightarrow$Depth}\tabularnewline
\midrule
\midrule 
Rectified Flow & 23.18/80.80 & 54.00/142.15 & 17.32/112.64 & \textcolor{brown}{64.47/75.58} & \textcolor{brown}{78.41/28.69} & \textcolor{brown}{79.20/35.53}\tabularnewline
\midrule 
UniDiffuser & 15.39/40.93 & 35.81/89.58 & 12.64/59.72 & \textcolor{brown}{39.62/38.44} & \textcolor{brown}{28.12/15.72} & \textcolor{brown}{38.39/23.41}\tabularnewline
\midrule 
iDR & \uline{10.72}\textbf{/}\uline{21.73} & \uline{21.64}\textbf{/}\uline{29.28} & \uline{7.25}/\uline{24.19} & \textbf{\textcolor{brown}{22.77/22.96}} & \textbf{\textcolor{brown}{17.88/8.63}} & \textbf{\textcolor{brown}{23.19/12.00}}\tabularnewline
\midrule 
dDR & \textbf{10.12}/\textbf{20.94} & \textbf{21.23}/\textbf{28.32} & \textbf{7.00}/\textbf{23.20} & \textcolor{brown}{\uline{26.73}}\textcolor{brown}{/}\textcolor{brown}{\uline{23.64}} & \textcolor{brown}{\uline{20.75}}\textcolor{brown}{/}\textcolor{brown}{\uline{9.42}} & \textcolor{brown}{\uline{24.91}}\textcolor{brown}{/}\textcolor{brown}{\uline{14.87}}\tabularnewline
\bottomrule
\end{tabular}}}
\par\end{centering}
\caption{FID scores on COCO-UMDT-Star. Translations without paired data are
marked in \textcolor{brown}{brown}. The best results are shown in
\textbf{bold}, and the second-best are \uline{underlined}.\label{tab:result_COCO-UMDT-Star}}
\end{table*}

\begin{table*}
\begin{centering}
{\resizebox{\textwidth}{!}{%
\par\end{centering}
\begin{centering}
\begin{tabular}{ccccccc}
\toprule 
\multirow{2}{*}{Method} & \multicolumn{6}{c}{FID$\downarrow$}\tabularnewline
\cmidrule{2-7} \cmidrule{3-7} \cmidrule{4-7} \cmidrule{5-7} \cmidrule{6-7} \cmidrule{7-7} 
 & Ske.$\leftrightarrow$Color & Seg.$\leftrightarrow$Color & \textcolor{brown}{Depth$\leftrightarrow$Color} & \textcolor{brown}{Ske.$\leftrightarrow$Seg.} & Ske.$\leftrightarrow$Depth & \textcolor{brown}{Seg.$\leftrightarrow$Depth}\tabularnewline
\midrule
\midrule 
Rectified Flow & 22.33/85.03 & 57.53/148.57 & \textcolor{brown}{27.49/146.27} & \textcolor{brown}{65.18/79.91} & 68.27/21.02 & \textcolor{brown}{94.55/42.40}\tabularnewline
\midrule 
UniDiffuser & 15.43/45.51 & 36.89/92.38 & \textcolor{brown}{14.40/68.64} & \textcolor{brown}{39.97/38.70} & 26.49/12.40 & \textcolor{brown}{39.85/24.51}\tabularnewline
\midrule 
iDR & \textbf{10.47/26.73} & \textbf{23.14/}\uline{38.60} & \textbf{\textcolor{brown}{8.55/39.66}} & \textbf{\textcolor{brown}{26.52/25.02}} & \textbf{14.73/7.54} & \textbf{\textcolor{brown}{26.08/14.15}}\tabularnewline
\midrule 
dDR & \uline{11.11}/\uline{28.39} & \uline{24.68}/\textbf{38.47} & \textcolor{brown}{\uline{10.69}}\textcolor{brown}{/}\textcolor{brown}{\uline{44.23}} & \textcolor{brown}{\uline{28.13}}\textcolor{brown}{/}\textcolor{brown}{\uline{26.94}} & \uline{14.92}/\uline{7.82} & \textcolor{brown}{\uline{30.98}}\textcolor{brown}{/}\textcolor{brown}{\uline{18.61}}\tabularnewline
\bottomrule
\end{tabular}}}
\par\end{centering}
\caption{FID scores on COCO-UMDT-Chain. Translations without paired data are
marked in \textcolor{brown}{brown}. The best results are shown in
\textbf{bold}, and the second-best are \uline{underlined}.\label{tab:result_COCO-UMDT-Chain}}
\vspace{-0.4em}
\end{table*}

\subsection{Results\label{subsec:main_result}}

We report the quantitative results for Shoes-UMDT and Faces-UMDT-Latent
in Table~\ref{tab:result_Shoes-UMDT}, for COCO-UMDT-Star in Table~\ref{tab:result_COCO-UMDT-Star},
and for Faces-UMDT-Pixel in Table~\ref{tab:result_Faces-UMDT} of
Appdx~\ref{subsec:Results-Faces-UMDT-Pixel}. iDR consistently outperforms
all baselines by a significant margin across all benchmarks, highlighting
the effectiveness of conditioning $\epsilon_{\theta}$ on both source
and target domain labels to guide translation. The weak performance
of StarGAN in central$\leftrightarrow$non-central translations ($X^{c}\leftrightarrow X^{k}$)
can be attributed to two main factors: (i) its relatively outdated
generator architecture, and (ii) its original design for unpaired
rather than paired translation. Despite being trained on aligned $\left(x^{c},x^{k}\right)$
pairs, Rectified Flow (RF) and UniDiffuser struggle to learn robust
$X^{c}\leftrightarrow X^{k}$ mappings. For RF, performance degrades
most noticeably when the target domain is diverse and high-variance
(e.g., Edge$\rightarrow$Shoe or Seg.$\rightarrow$Face), as its deterministic
formulation cannot capture the stochasticity of $p\left(x^{c}|x^{k}\right)$
and $p\left(x^{k}|x^{c}\right)$. For UniDiffuser, the repeated substitution
of missing domains with Gaussian noise during training and inference
likely undermines its ability to accurately modeling the joint distribution
over all domains.

dDR shows slight decreases in performance for translation tasks without
paired data compared to iDR, yet still outperforms all baselines significantly.
The performance drop can be attributed to imperfections in our refinement
procedure and to bias introduced by the Monte Carlo approximation
inside the logarithm in Eq.~\ref{eq:log_MC_estimate_1}. Importantly,
dDR reduces the number of sampling steps for non-central translations
by half relative to iDR--a substantial efficiency gain, given that
iDR typically requires hundreds to thousands of steps for cross-domain
translation. As shown in Fig.~\ref{fig:Qualitative-results}, iDR
and dDR produce higher-quality samples compared to the baselines.

We further observe consistent behavior on COCO-UMDT-Chain in Table~\ref{tab:result_COCO-UMDT-Chain}
compared to COCO-UMDT-Star: dDR improves efficiency by 2-3 times for
translations without paired data while incurring only marginal performance
degradation compared to iDR. This empirically validate the generalizability
of our learning strategy for UMDT beyond star-shaped structures.

\subsection{Ablation Studies}

Owing to space limitations, we conduct comprehensive ablations to
quantify the effect of key hyperparameter choices and report at Appdx.~\ref{subsec:Ablation-Studies}.
Specifically, we (i) evaluate Tweedie refinement by varying the number
of refinement steps $n\in\left\{ 0,1,3,5\right\} $ (Appdx.~\ref{subsec:abla_impact_n});
(ii) study the effect of the rehearsal coefficient $\lambda_{2}\in\left\{ 0,0.3,1,3\right\} $
in the loss $\Loss_{\text{final}}$ during finetuning (Appdx.~\ref{subsec:abla_impact_lambda});
compare dDR trained from scratch versus finetuned from a pretrained
iDR (Appdx.~\ref{subsec:Training-from-scratch}); (iv) compare the
performances of iDR and dDR across different numbers of sampling steps
(Appdx.~\ref{subsec:abla_different_NFE}); and (v) benchmark standard
DR against several bridge-based DR variants on UMDT (Appdx.~\ref{subsec:Comparison-of-Standard}).

%% file: relate.tex
Due to space constraints, we primarily review diffusion-based approaches
for multi-domain translation (MDT), with other methods discussed in
Appdx.~\ref{sec:Additional-related-work}. Building on the success
of Stable Diffusion (SD) \cite{rombach2022high} in text-to-image
generation, many works adapt SD to MDT by finetuning it to condition
on additional modalities such as edges, segmentation maps, depth,
or poses \cite{zhang2023adding,huang2023composer,Mou2024}. Versatile
Diffusion \cite{xu2023versatile} extends this idea with a modular
cross-modal architecture, but its model size scales linearly with
the number of domains and experiments are limited to text and image.
CoDi \cite{tang2023any} adopts a two-stage framework: first, modality-specific
encoders are trained to align samples into a shared latent space;
second, separate latent diffusion models are trained for each modality
conditioned on encoder outputs and noisy paired samples. While this
supports any-to-any generation, it requires contrastive pretraining
and multiple diffusion models that scale linearly with domain count.
UniDiffuser \cite{bao2023one} instead models the joint distribution
of all domains using a single transformer-based noise prediction network
that treats domain samples as tokens, with independent noise schedules
per domain. This design supports arbitrary modalities but requires
fully aligned tuples across domains and long training times. Moreover,
UniDiffuser must process all domains jointly, making generation costly
when only a subset of domains is desired. One Diffusion~\cite{le2025one}
follows a similar design but replaces the noise prediction network
with a velocity network trained via flow matching \cite{liu2022flow}
and introduces a different transformer architecture \cite{zhuo2024lumina}.
Like UniDiffuser, it depends on fully aligned tuples where samples
of missing modalities are synthesized. OmniFlow \cite{li2025omniflow}
is another flow-based model for multi-modal generation like One-Diffusion.
However, instead of training from scratch, it extends from SD3~\cite{patrick2024scaling}.

In summary, most diffusion-based MDT methods either rely on contrastive
learning or synthetic fully aligned data, suffer from linear growth
in model size, or require processing all domains simultaneously. By
contrast, our method trains directly from domain pairs, avoids model-size
scaling with domain count, and flexibly adjusts sampling cost to the
desired number of output domains. Although not the primary focus of
this work, Diffusion Routers also enable generation of a single target
domain from multiple source domains by combining the scores of models
conditioned on these sources. This capability makes our method adaptable
for any-to-any generation.

%% file: discuss.tex
We introduced the universal multi-domain translation (UMDT) problem,
which seeks to learn mappings between any pair of $K$ domains using
only $K-1$ paired datasets with a central domain. The main challenge
lies in learning non-central$\leftrightarrow$non-central translations
(NNTs), where training pairs are unavailable. To tackle this, we proposed
Diffusion Router (DR), which supports both indirect NNTs through the
central domain and direct NNTs. The direct variant can be obtained
either by fine-tuning from the indirect version or by training from
scratch. We introduced novel variational-bound objective and conditional
sampling method for learning the direct variant. Empirical evaluations
on three newly constructed UMDT datasets demonstrated that our method
consistently outperforms existing baselines. In future work, we aim
to extend DR to large-scale multimodal generation across image, text,
and audio, particularly addressing scenarios where paired datasets
(e.g., image$\leftrightarrow$audio) are scarce in practice.

%% file: appdx.tex
\section{Theoretical Results}

\subsection{Derivation of Eq.~\ref{eq:orig_KL_div}\label{subsec:Mathematical-Derivation-of}}

The detailed derivation of Eq.~\ref{eq:orig_KL_div} is given below:
\begin{align}
 & \mathbb{E}_{p\left(x^{i}\right)}\left[D_{KL}\left(p\left(x^{j}|x^{i}\right)||p_{\theta}\left(x^{j}\mid x^{i}\right)\right)\right]\nonumber \\
=\  & \int p\left(x^{i}\right)\left[\int p\left(x^{j}|x^{i}\right)\left(\log p\left(x^{j}|x^{i}\right)-\log p_{\theta}\left(x^{j}|x^{i}\right)\right)dx^{j}\right]dx^{i}\\
=\  & \int p\left(x^{i}\right)\int\left(\int p\left(x^{j}|x^{c}\right)p\left(x^{c}|x^{i}\right)dx^{c}\right)\left(\log p\left(x^{j}|x^{i}\right)-\log p_{\theta}\left(x^{j}|x^{i}\right)\right)dx^{j}dx^{i}\\
=\  & \int\int\int p\left(x^{i}\right)p\left(x^{c}|x^{i}\right)p\left(x^{j}|x^{c}\right)\left(\log p\left(x^{j}|x^{i}\right)-\log p_{\theta}\left(x^{j}|x^{i}\right)\right)dx^{j}dx^{c}dx^{i}\\
=\  & \int p\left(x^{i},x^{c}\right)\int p\left(x^{j}|x^{c}\right)\left(\log p\left(x^{j}|x^{i}\right)-\log p_{\theta}\left(x^{j}|x^{i}\right)\right)dx^{j}dx^{c}dx^{i}\\
=\  & \Expect_{p\left(x^{i},x^{c}\right)}\Expect_{p\left(x^{j}|x^{c}\right)}\left[\log p\left(x^{j}|x^{i}\right)-\log p_{\theta}\left(x^{j}|x^{i}\right)\right]\\
=\  & \Expect_{p\left(x^{i},x^{c}\right)}\Expect_{p\left(x^{j}|x^{c}\right)}\left[\log\Expect_{p\left(x'^{c}|x^{i}\right)}\left[p\left(x^{j}|x'^{c}\right)\right]-\log p_{\theta}\left(x^{j}|x^{i}\right)\right]
\end{align}
Note that in our derivation, $p\left(x^{j}|x^{i}\right)$ is substituted
with its expression in described in Eq.~\ref{eq:cond_distr}.

\subsection{Derivation of the Variational Upper Bound in Eq.~\ref{eq:VB_path_KL_div}\label{subsec:Variational-Upper-Bound}}

The detailed derivation of the variational upper bound in Eq.~\ref{eq:VB_path_KL_div}
is as follows:

\begin{align}
 & \mathbb{E}_{\left(x^{i},x^{c}\right)\sim\mathcal{D}_{i,c}}\left[D_{KL}\left(p_{\text{ref}}\left(x^{j}|x^{c}\right)||p_{\theta}\left(x^{j}|x^{i}\right)\right)\right]\nonumber \\
\le\  & \mathbb{E}_{\left(x^{i},x^{c}\right)\sim\mathcal{D}_{i,c}}\left[D_{KL}\left(p_{\text{ref}}\left(x_{0:T}^{j}|x^{c}\right)||p_{\theta}\left(x_{0:T}^{j}|x^{i}\right)\right)\right]\label{eq:VUB_1}\\
=\  & \mathbb{E}_{\left(x^{i},x^{c}\right)\sim\mathcal{D}_{i,c}}\left[\int p_{\text{ref}}\left(x_{0:T}^{j}|x^{c}\right)\log\left(\frac{p_{\text{ref}}\left(x_{0:T}^{j}|x^{c}\right)}{p_{\theta}\left(x_{0:T}^{j}|x^{i}\right)}\right)dx_{0:T}^{j}\right]\\
=\  & \mathbb{E}_{\left(x^{i},x^{c}\right)\sim\mathcal{D}_{i,c}}\left[\int p_{\text{ref}}\left(x_{0:T}^{j}|x^{c}\right)\log\left(\frac{p\left(x_{T}^{j}|x^{c}\right)\prod_{t=1}^{T}p_{\text{ref}}\left(x_{t-1}^{j}|x_{t}^{j},x^{c}\right)}{p\left(x_{T}^{j}|x^{i}\right)\prod_{t=1}^{T}p_{\theta}\left(x_{t-1}^{j}|x_{t}^{j},x^{i}\right)}\right)dx_{0:T}^{j}\right]\\
=\  & \mathbb{E}_{\left(x^{i},x^{c}\right)\sim\mathcal{D}_{i,c}}\left[\mathbb{E}_{p_{\text{ref}}\left(x_{0:T}^{j}|x^{c}\right)}\left[\sum_{t=1}^{T}\log\left(\frac{p_{\text{ref}}\left(x_{t-1}^{j}|x_{t}^{j},x^{c}\right)}{p_{\theta}\left(x_{t-1}^{j}|x_{t}^{j},x^{i}\right)}\right)\right]\right]+\text{const}\\
=\  & \mathbb{E}_{\left(x^{i},x^{c}\right)\sim\mathcal{D}_{i,c}}\left[\sum_{t=1}^{T}\mathbb{E}_{p_{\text{ref}}\left(x_{t}^{j}\mid x^{c}\right)}\mathbb{E}_{p_{\text{ref}}\left(x_{t-1}^{j}\mid x_{t}^{j},x^{c}\right)}\left[\log\left(\frac{p_{\text{ref}}\left(x_{t-1}^{j}|x_{t}^{j},x^{c}\right)}{p_{\theta}\left(x_{t-1}^{j}|x_{t}^{j},x^{i}\right)}\right)\right]\right]+\text{const}\\
=\  & \mathbb{E}_{\left(x^{i},x^{c}\right)\sim\mathcal{D}_{i,c}}\left[\sum_{t=1}^{T}\mathbb{E}_{p_{\text{ref}}\left(x_{t}^{j}\mid x^{c}\right)}D_{KL}\left(p_{\text{ref}}\left(x_{t-1}^{j}|x_{t}^{j},x^{c}\right)||p_{\theta}\left(x_{t-1}^{j}|x_{t}^{j},x^{i}\right)\right)\right]+\text{const}
\end{align}
Here, the inequality in Eq.~\ref{eq:VUB_1} is a variant of the data
processing inequality.

\section{Additional Preliminaries}

\subsection{Diffusion Bridges\label{subsec:prelim-diffusion-bridges}}

Diffusion bridges \cite{LiuVHTNA23,albergo2023stochastic,LiuW0l23,LiX0L23,zhou2024denoising,Kieu2025}
offer an alternative approach to modeling conditional distributions
$p\left(x|y\right)$ where the generation process starts from the
observation $y$ rather than from standard Gaussian noise, as in traditional
diffusion models. Specifically, the corruption process is characterized
by the marginal distribution $p\left(x_{t}|x,y\right)$, representing
a stochastic trajectory between endpoints $x$ and $y$. This distribution
is assumed to be Gaussian: $p\left(x_{t}|x,y\right)=\mathcal{N}\left(\alpha_{t}x+\beta_{t}y,\sigma_{t}^{2}\mathrm{I}\right)$,
which allows $x_{t}$ to be directly sampled from a training pair
$\left(x,y\right)$ using:
\begin{equation}
x_{t}=\alpha_{t}x_{0}+\beta_{t}y+\sigma_{t}\epsilon,\label{eq:bridge_fwd_xt}
\end{equation}

\noindent where $\epsilon\sim\mathcal{N}\left(0,\mathrm{I}\right)$;
$\alpha_{t}$, $\beta_{t}$, $\sigma_{t}$ are time-dependent coefficients
satisfying boundary conditions: $\alpha_{0}=\beta_{T}=1$ and $\alpha_{T}=\beta_{0}=\sigma_{0}=\sigma_{T}=0$.

The transition distribution $p_{\theta}\left(x_{t-1}|x_{t},y\right)$
for generating $x$ from $y$ (with $x_{T}\equiv y$) is modeled as
$\mathcal{N}\left(\mu_{\theta,t,t-1}\left(x_{t},y\right),\delta_{t-1|t}^{2}\frac{\sigma_{t-1}^{2}}{\sigma_{t}^{2}}\mathrm{I}\right)$
where the mean is given by \cite{Kieu2025}:
\begin{align}
\mu_{\theta,t,t-1}\left(x_{t},y\right) & =\frac{\alpha_{t-1}}{\alpha_{t}}x_{t}+\left(\beta_{t-1}-\frac{\beta_{t}\alpha_{t-1}}{\alpha_{t}}\right)y+\left(\frac{\sigma_{t-1}\sqrt{\sigma_{t}^{2}-\delta_{t-1|t}^{2}}}{\sigma_{t}}-\frac{\sigma_{t}\alpha_{t-1}}{\alpha_{t}}\right)\epsilon_{\theta}\left(x_{t},t,y\right)\label{eq:bridge_bck_mean}
\end{align}

\noindent Here, $\delta_{t-1|t}\in[0,\text{\ensuremath{\sigma_{t}}})$
controls the sampling variance and is typically defined as $\delta_{t-1|t}:=\sqrt{\eta\left(\sigma_{t}^{2}-\sigma_{t-1}^{2}\frac{\alpha_{t-1}^{2}}{\alpha_{t}^{2}}\right)}$
with $\eta\in\left[0,1\right]$ in the case of Brownian bridges \cite{LiX0L23}. 

\noindent The noise prediction network $\epsilon_{\theta}$ in Eq.~\ref{eq:bridge_bck_mean}
is trained by minimizing the noise matching loss:
\begin{equation}
\mathcal{L}\left(\theta\right)=\mathbb{E}_{\left(x,y\right),t,\epsilon}\left[\mathbb{E}_{x_{t}}\left[\left\Vert \epsilon_{\theta}\left(x_{t},t,y\right)-\epsilon\right\Vert _{2}^{2}\right]\right],\label{eq:bridge_loss}
\end{equation}

\noindent where $\left(x,y\right)\sim p\left(x,y\right)$, $t\sim\mathcal{U}\left(1,T\right)$,
$\epsilon\sim\mathcal{N}\left(0,\mathrm{I}\right)$.

\noindent Bidirectional diffusion bridges \cite{Kieu2025} extend
this framework to jointly model both $p\left(x|y\right)$ and $p\left(y|x\right)$
using a single shared noise prediction network.

\section{Additional Related Work\label{sec:Additional-related-work}}

Multi-domain translation (MDT) methods can be classified according
to the underlying generative models. This section focuses on reviewing
traditional VAE-based and GAN-based approaches.

\subsection{VAE-based methods}

VAE-based methods generally aim to learn latent representations that
facilitate translation across domains or modalities. JMVAE \cite{suzuki2016joint}
captures shared representations with a joint encoder $q_{\theta}\left(z\mid x^{1},x^{2}\right)$
and handles missing modalities at test time by aligning unimodal encoders
$q_{\theta}\left(z\mid x^{1}\right)$ and $q_{\theta}\left(z\mid x^{2}\right)$
with the joint encoder through KL divergence minimization. TELBO \cite{vedantam2018generative}
adopts a similar encoder design but differs in its training strategy.
Instead of jointly optimizing all encoders, it first trains the joint
encoder and then fits the unimodal encoders while keeping the joint
encoder\textquoteright s parameters fixed. MFM \cite{tsai2019learning}
factorizes the multimodal latent space into shared discriminative
and modality-specific generative factors, enabling inference of missing
modalities at test time from observed modalities. While these methods
are sufficient for two modalities, they do not generalize to the truly
multi-modal case.

More recently, MVAE \cite{wu2018multimodal} innovatively factorize
the joint encoder into a product of experts (PoE), i.e., $q_{\Phi}\left(z\mid x^{1},x^{2}\right)=q_{\phi}\left(z\mid x^{1}\right)q_{\phi}\left(z\mid x^{2}\right)p\left(z\right)$,
a notable advance that scales to multiple domains by training with
randomly masked modalities and seamless inference when any modality
is missing at test time. Alternately, MMVAE \cite{shi2019variational}
constructs the joint encoder as a mixture of experts (MoE) of unimodal
encoders, alleviating the precision\nobreakdash-miscalibration issues
inherent to PoE. Despite strong results in conventional MDT, both
MVAE and MMVAE assume fully aligned tuples across domains, a requirement
rarely satisfied in practice. In contrast, our method trains directly
on domain pairs, removing the need for fully aligned triplets and
improving practicality in real-world settings.

\subsection{GAN-based methods}

\begin{table*}
\begin{centering}
{\resizebox{\textwidth}{!}{%
\par\end{centering}
\begin{centering}
\begin{tabular}{ccccccc}
\toprule 
\multirow{3}{*}{Method} & \multicolumn{6}{c}{LPIPS $\downarrow$}\tabularnewline
\cmidrule{2-7} \cmidrule{3-7} \cmidrule{4-7} \cmidrule{5-7} \cmidrule{6-7} \cmidrule{7-7} 
 & \multicolumn{3}{c}{Shoes-UMDT} & \multicolumn{3}{c}{Faces-UMDT-Latent}\tabularnewline
\cmidrule{2-7} \cmidrule{3-7} \cmidrule{4-7} \cmidrule{5-7} \cmidrule{6-7} \cmidrule{7-7} 
 & Edge$\leftrightarrow$Shoe & Gray.$\leftrightarrow$Shoe & \textcolor{brown}{Edge$\leftrightarrow$Gray.} & Ske.$\leftrightarrow$Face & Seg.$\leftrightarrow$Face & \textcolor{brown}{Ske.$\leftrightarrow$Seg.}\tabularnewline
\midrule
\midrule 
StarGAN & 0.128/0.223 & 0.095/0.214 & \textcolor{brown}{0.191/0.144} & - & - & \textcolor{brown}{-}\tabularnewline
Rectified Flow & \uline{0.063}/0.175 & 0.012/0.146 & \textcolor{brown}{0.138/}\textcolor{brown}{\uline{0.083}} & 0.278/0.560 & 0.165/0.548 & \textcolor{brown}{0.419/0.274}\tabularnewline
Unidiffuser & 0.066/0.170 & 0.019/0.091 & \textcolor{brown}{0.090/0.091} & 0.263/0.483 & 0.165/0.511 & \textcolor{brown}{0.393/0.223}\tabularnewline
\midrule 
iDR & \textbf{0.050}/\uline{0.129} & \textbf{0.003}/\uline{0.069} & \textbf{\textcolor{brown}{0.069}}\textcolor{brown}{/}\textbf{\textcolor{brown}{0.058}} & \textbf{0.221}/\textbf{0.427} & \uline{0.129}/\uline{0.471} & \textbf{\textcolor{brown}{0.377}}\textcolor{brown}{/}\textbf{\textcolor{brown}{0.177}}\tabularnewline
dDR & \textbf{0.050}/\textbf{0.128} & \uline{0.004}/\textbf{0.063} & \textcolor{brown}{\uline{0.077}}\textcolor{brown}{/0.090} & \uline{0.222}/\uline{0.428} & \textbf{0.126}/\textbf{0.466} & \textcolor{brown}{\uline{0.392}}\textcolor{brown}{/}\textcolor{brown}{\uline{0.192}}\tabularnewline
\bottomrule
\end{tabular}}}
\par\end{centering}
\caption{LPIPS scores of our method and baselines on Shoes-UMDT and Faces-UMDT-Latent.
Translations without paired data are marked in \textcolor{brown}{brown}.
The best results are shown in \textbf{bold}, and the second-best are
\uline{underlined}.\label{tab:LPIPS-on-Shoes-UMDT}}
\end{table*}

Another line of research uses GAN \cite{goodfellow2014generative}
for MDT. Pix2pix \cite{isola2017image} uses paired datasets to train
a conditional GAN with a reconstruction loss, aligning outputs with
ground-truth targets while encouraging realism. CoGAN \cite{liu2016coupled}
instead learns a joint distribution across domains by sharing weights
in high-level layers, enabling related samples across domains without
aligned pairs. However, weight sharing constrains model design and
limits scalability to high-resolution images. A popular approach for
MDT is to leverage cycle-consistency constraint when training cross-domain
translators \cite{zhu2017unpaired,kim2017learning,yi2017dualgan,liu2017unsupervised},
enforcing a meaningful relation between the input and the translated
image. Nevertheless, these GAN-based methods do not scale gracefully
with more domains, leading to higher computational cost.

Notably, StarGAN~\cite{choi2018stargan} introduces a unified GAN
framework for MDT that uses a single generator and discriminator with
an auxiliary domain classifier during training. Its successor, StarGAN-2~\cite{choi2020stargan},
injects continuous style codes via AdaIN~\cite{huang2017arbitrary}
to produce diverse outputs. Like StarGAN, UFDN \cite{liu2018unified}
also offers a unified MDT model and and extends the framework to disentangle
domain-invariant features. More recently, MultimodalGAN \cite{Zhu2024}
proposes an MDT framework trained on fully aligned tuples across domains.

Our work is related to StarGAN in that both introduce frameworks for
MDT, but differs in supervision and translation. We exploit paired
datasets for training and perform indirect translations between domains
lacking training pairs. In contrast, StarGAN treats such translations
as an unpaired translations. We empirically show that leveraging paired
supervision leads to more faithful input--output mappings and superior
performance.

\section{Additional Experiment}

\subsection{Additional Implementation Details\label{subsec:Implementation-Detail}}

\subsubsection{Diffusion Router}

\begin{figure}
\centering{}\includegraphics[width=0.95\textwidth]{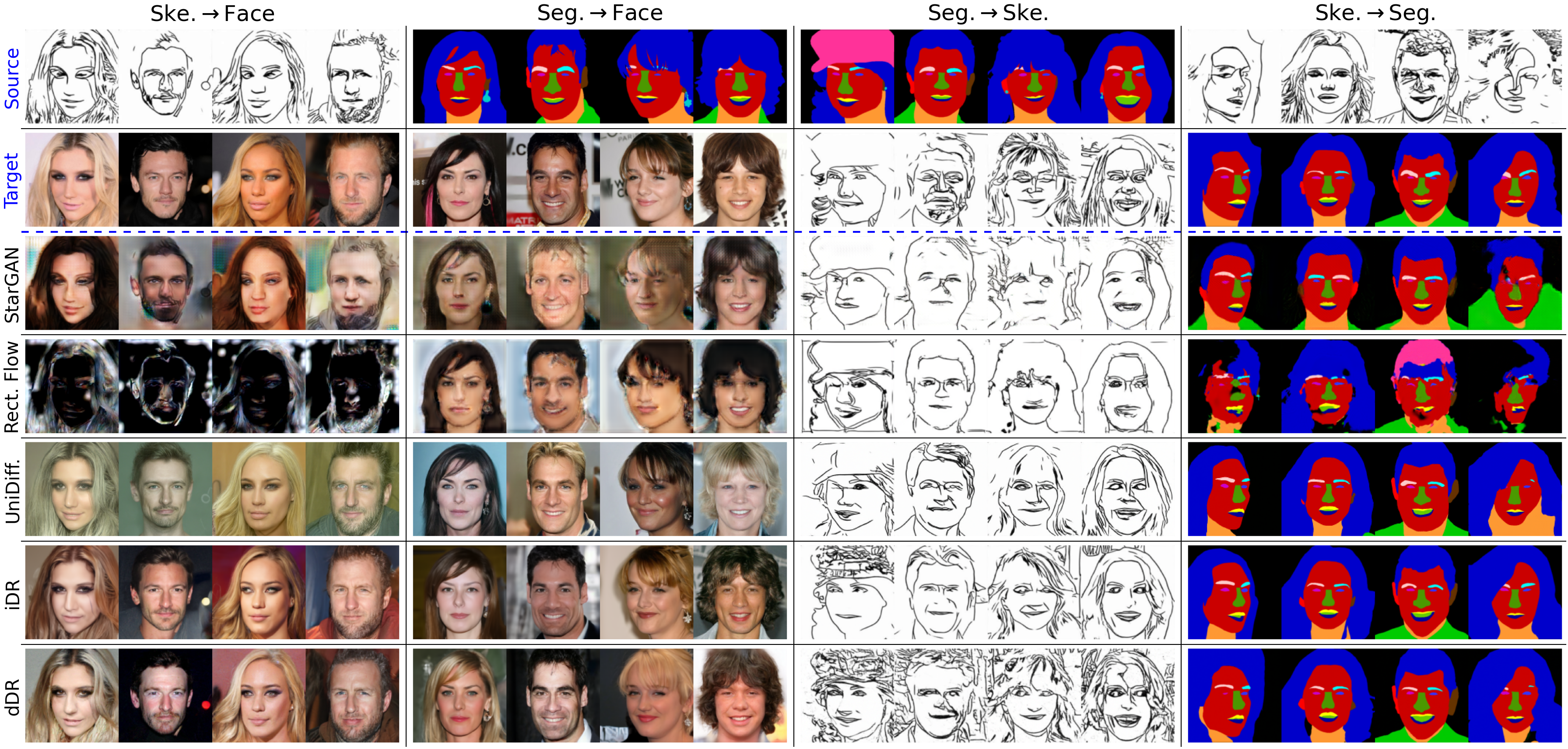}\caption{Qualitative results of our method and baselines on Faces-UMDT-Pixel.\label{fig:result_faces_pixel}}
\end{figure}

We adopt the standard Diffusion Router variant with DDPM \cite{ho2020denoising}
as the underlying diffusion process. A comparison with the bridge-based
variant is provided in Appdx.~\ref{subsec:Comparison-of-Standard}.
The noise prediction network $\epsilon_{\theta}$ has a U-Net architecture
following \cite{dhariwal2021diffusion}. For Shoes-UMDT (64$\times$64)
and Faces-UMDT-Pixel (128$\times$128), $\epsilon_{\theta}$ operates
directly on raw images with 128 and 256 base channels, respectively.
For Faces-UMDT-Latent, COCO-UMDT-Star, and COCO-UMDT-Chain, the 256$\times$256
images are first encoded into 32\texttimes 32\texttimes 4 latents
using a pretrained VAE \cite{rombach2022high}, which are then processed
by $\epsilon_{\theta}$ with 128 base channels. To ensure efficient
training within a single H100-80GB GPU, we use a batch size of 128
for the 128-channel U-Net variant and 32 for the 256-channel variant.
iDR is trained using AdamW \cite{Loshchilov2019} with a learning
rate of 1e-4, $\beta_{1}=\beta_{2}=0.9$ and 3000 warm-up steps across
all datasets. The number of training steps is 250k for Shoes-UMDT
and Faces-UMDT, and 500k for COCO-UMDT. Training takes roughly three
to four days. dDR is finetuned from a pretrained iDR with a reduced
learning rate of 5e-5, loss coefficients $\lambda_{1}=1$ and $\lambda_{2}=1$,
and 5 refinement steps. The number of finetuning steps is 100k for
Shoes-UMDT and Faces-UMDT, and 150k for COCO-UMDT. We run DDIM \cite{song2020denoising}
with 1000 steps to translate from one domain to another.

\subsubsection{Baselines}

We implement StarGAN using its \href{https://github.com/yunjey/stargan}{official repository}.
While more recent GAN-based approaches for MDT exist, such as MultimodalGAN
\cite{Zhu2024}, we exclude them from our baselines due to the lack
of publicly available code and instead discuss them in Related Work. 

UniDiffuser \cite{bao2023one} is designed to model noisy samples
across all domains at random noise levels. Since our training data
provide only domain pairs rather than full tuples, we substitute missing
domains with Gaussian noise and set the corresponding domain-specific
time step to 999 during training. Rectified Flow \cite{liu2022flow},
which natively supports only two-domain translation with each domain
as a boundary distribution, is adapted to the multi-domain setting
by conditioning the velocity network on both source and target domain
labels, similar to our approach. For fairness, we implement UniDiffuser
and Recified Flow using the same architecture and settings as our
method.

\subsection{Additional Results}

\subsubsection{Shoes-UMDT and Faces-UMDT-Latent}

Table~\ref{tab:LPIPS-on-Shoes-UMDT} shows the LPIPS scores for our
method and the baselines. Overall, iDR and dDR consistenly outperform
the baselines across all translation tasks. On tasks with paired supervision,
both methods achieve comparable results. For translations without
paired data, dDR exhibits a slight performance drop relative to iDR,
but supports direct non-central translations. These observations align
with the FID results presented in the main text.

\subsubsection{Faces-UMDT-Pixel\label{subsec:Results-Faces-UMDT-Pixel}}

\begin{table*}
\begin{centering}
{\resizebox{\textwidth}{!}{%
\par\end{centering}
\begin{centering}
\begin{tabular}{ccccccc}
\toprule 
\multirow{2}{*}{Method} & \multicolumn{3}{c}{FID$\downarrow$} & \multicolumn{3}{c}{LPIPS$\downarrow$}\tabularnewline
\cmidrule{2-7} \cmidrule{3-7} \cmidrule{4-7} \cmidrule{5-7} \cmidrule{6-7} \cmidrule{7-7} 
 & Ske.$\leftrightarrow$Face & Seg.$\leftrightarrow$Face & \textcolor{brown}{Ske.$\leftrightarrow$Seg.} & Ske.$\leftrightarrow$Face & Seg.$\leftrightarrow$Face & \textcolor{brown}{Ske.$\leftrightarrow$Seg.}\tabularnewline
\midrule
\midrule 
StarGAN & 38.16/91.67 & 32.98/75.41 & \textcolor{brown}{50.17/76.40} & 0.280/0.479 & 0.282/0.471 & \textcolor{brown}{0.408/0.364}\tabularnewline
Rectified Flow & 9.88/261.96 & 19.54/122.87 & \textcolor{brown}{45.01/109.37} & 0.160/0.635 & 0.137/0.464 & \textcolor{brown}{\uline{0.362}}\textcolor{brown}{/0.398}\tabularnewline
Unidiffuser & 18.49/36.99 & 14.03/25.36 & \textcolor{brown}{36.89/15.79} & 0.181/0.548 & 0.125/0.442 & \textcolor{brown}{0.367/0.191}\tabularnewline
\midrule 
iDR & \uline{9.25}/\uline{13.32} & \uline{4.1}2/\textbf{10.54} & \textbf{\textcolor{brown}{13.51}}\textcolor{brown}{/}\textbf{\textcolor{brown}{3.88}} & \uline{0.159}/\textbf{0.329} & \textbf{0.101}/\textbf{0.412} & \textbf{\textcolor{brown}{0.361}}\textcolor{brown}{/}\textbf{\textcolor{brown}{0.146}}\tabularnewline
dDR & \textbf{9.02}/\textbf{12.81} & \textbf{3.57}/\uline{12.88} & \textcolor{brown}{\uline{28.46}}\textcolor{brown}{/}\textcolor{brown}{\uline{3.91}} & \textbf{0.139}/\uline{0.347} & \uline{0.106}/\uline{0.421} & \textcolor{brown}{0.422/}\textcolor{brown}{\uline{0.167}}\tabularnewline
\bottomrule
\end{tabular}}}
\par\end{centering}
\caption{Results on Faces-UMDT-Pixel of our method and baselines. Translations
without paired data are marked in \textcolor{brown}{brown}. The best
results are shown in \textbf{bold}, and the second-best are \uline{underlined}.\label{tab:result_Faces-UMDT}}
\end{table*}

For completeness, we compare iDR and dDR against baselines on Faces-UMDT-Pixel
(see Table \ref{tab:result_Faces-UMDT}). The results from Table \ref{tab:result_Faces-UMDT}
mirror the trends in our main experiments at Section~\ref{subsec:main_result},
with iDR consistently surpassing all baselines and showing the largest
gains on high uncertainty translations (e.g., Ske.$\to$Face and Seg.$\to$Face).
For Rectified Flow (RF), we observe poor generalization on the test
set, particularly on Ske.$\to$Face, yielding underperforming results.
This underperformance likely arises because training uses FFHQ for
Ske.$\to$Face while testing use CelebA-HQ for Ske.$\to$Face, causing
a mismatch between the training and test distributions and encourging
model's generalization, where RF fails.

In comparison, dDR performs comparably to iDR on most translations
while maintaining a clear margin over the baselines. We attribute
dDR\textquoteright s lower results on Seg.$\to$Ske. to FID\textquoteright s
sensitivity for sketches: background differences and minor detail
changes can depress the score even when outputs are visually similar
to the targets (see Fig.~\ref{fig:result_faces_pixel}). These results
empirically indicate that iDR and dDR can generalize to high-resolution
data.

\subsection{Ablation Studies\label{subsec:Ablation-Studies}}

Unless otherwise specified, ablation studies are conducted on Faces-UMDT-Latent
to evaluate the impact of different design choices. To reduce both
training and sampling costs, we substitute the large U-Net used in
the main experiments (304.8M parameters) with a smaller variant (32.3M
parameters).

\begin{figure*}[htpb]

\noindent %
\begin{minipage}[t]{0.45\textwidth}%
\begin{center}
\begin{table}[H]
\begin{centering}
{\resizebox{1.0\textwidth}{!}{
\begin{tabular}{ccccc}
\hline 
\multirow{2}{*}{} & \multirow{2}{*}{$n$} & \multicolumn{3}{c}{FID$\downarrow$}\tabularnewline
\cline{3-5} \cline{4-5} \cline{5-5} 
 &  & Ske.$\leftrightarrow$Face & Seg.$\leftrightarrow$Face & \textcolor{brown}{Ske.$\leftrightarrow$Seg.}\tabularnewline
\hline 
\hline 
iDR & - & 14.21/39.06 & 10.18/24.95 & \textcolor{brown}{20.82/10.85}\tabularnewline
\hline 
\multirow{4}{*}{dDR} & 0 & 16.11/39.82 & 13.75/25.77 & \textcolor{brown}{55.30/13.35}\tabularnewline
 & 1 & 14.14/40.61 & 11.77/23.65 & \textcolor{brown}{37.45/13.13}\tabularnewline
 & 3 & 13.73/41.14 & 11.58/25.26 & \textcolor{brown}{26.77/10.96}\tabularnewline
 & 5 & 13.52/37.26 & 11.52/22.73 & \textcolor{brown}{26.27/11.37}\tabularnewline
\hline 
\end{tabular}}}
\par\end{centering}
\caption{FID scores of dDR finetuned on Faces-UMDT-Latent w.r.t. different
number of refinement steps $n$.\label{tab:abl_refine_steps}}
\end{table}
\par\end{center}%
\end{minipage}\hfill{}%
\begin{minipage}[t]{0.52\textwidth}%
\begin{center}
\begin{table}[H]
\centering{}{\resizebox{1.0\textwidth}{!}{
\begin{tabular}{ccccc}
\hline 
\multirow{2}{*}{} & \multirow{2}{*}{$\lambda_{2}$} & \multicolumn{3}{c}{FID$\downarrow$}\tabularnewline
\cline{3-5} \cline{4-5} \cline{5-5} 
 &  & Ske.$\leftrightarrow$Face & Seg.$\leftrightarrow$Face & \textcolor{brown}{Ske.$\leftrightarrow$Seg.}\tabularnewline
\hline 
\hline 
iDR & - & 14.21/39.06 & 10.18/24.95 & \textcolor{brown}{20.82/10.85}\tabularnewline
\hline 
\multirow{4}{*}{dDR} & 0 & 405.12/306.61 & 355.78/307.15 & \textcolor{brown}{403.19/328.75}\tabularnewline
 & 0.3 & 19.44/45.94 & 17.01/30.90 & \textcolor{brown}{52.31/13.18}\tabularnewline
 & 1 & 16.11/39.82 & 13.75/25.77 & \textcolor{brown}{55.30/13.35}\tabularnewline
 & 3 & 14.47/39.38 & 10.19/25.50 & \textcolor{brown}{82.77/13.23}\tabularnewline
\hline 
\end{tabular}}}\caption{FID scores of dDR finetuned on Faces-UMDT-Latent w.r.t. different
values of $\lambda_{2}$. Tweedie refinement is not applied in this
case (i.e, $n=0$).\label{tab:ablation_lambda}}
\end{table}
\par\end{center}%
\end{minipage}

\noindent \end{figure*}

\subsubsection{Impact of the number of refinement steps $n$\label{subsec:abla_impact_n}}

\begin{figure}
\begin{centering}
\begin{tabular}{ccc}
\includegraphics[width=0.34\textwidth]{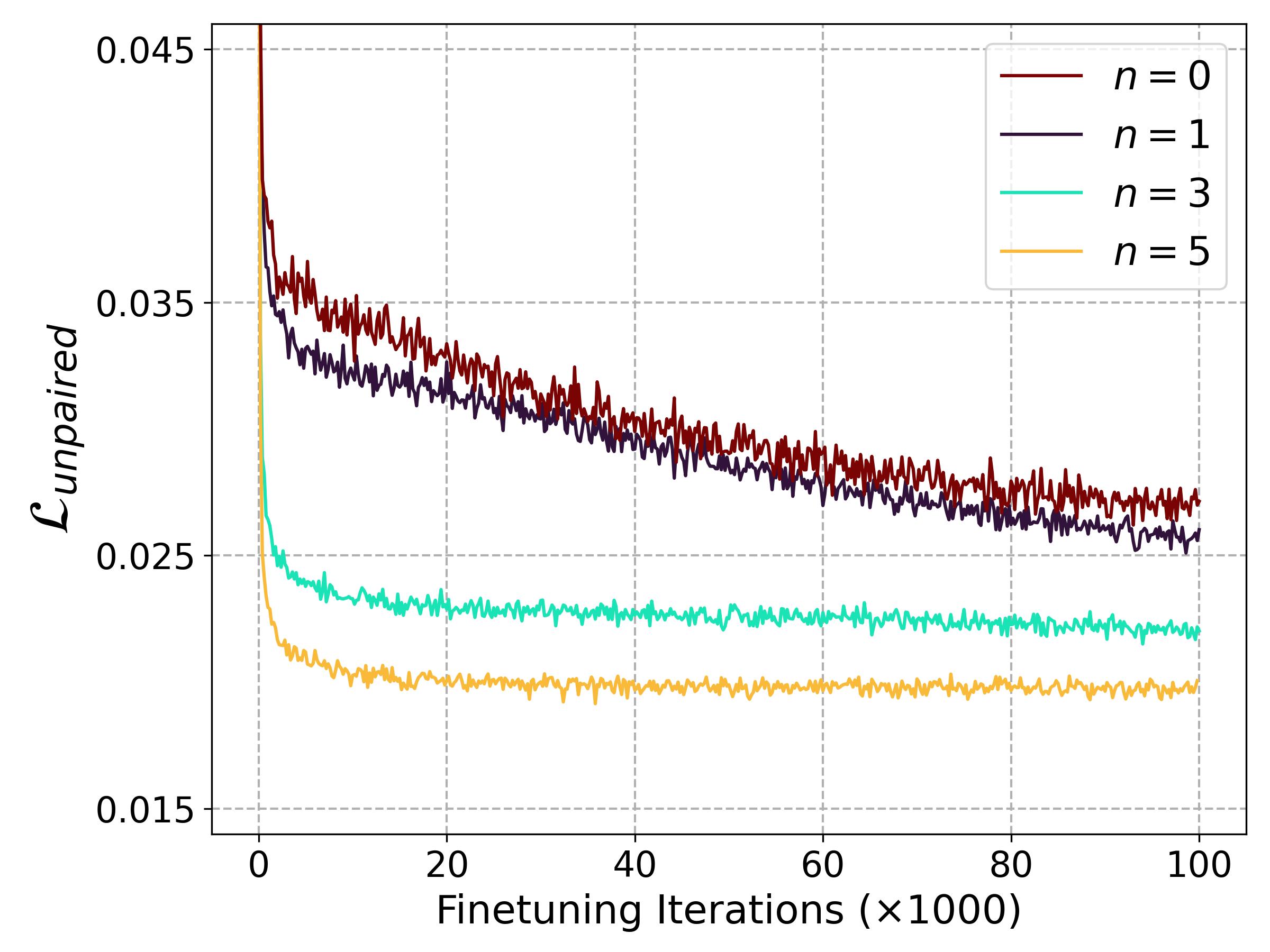} &  & \includegraphics[width=0.34\textwidth]{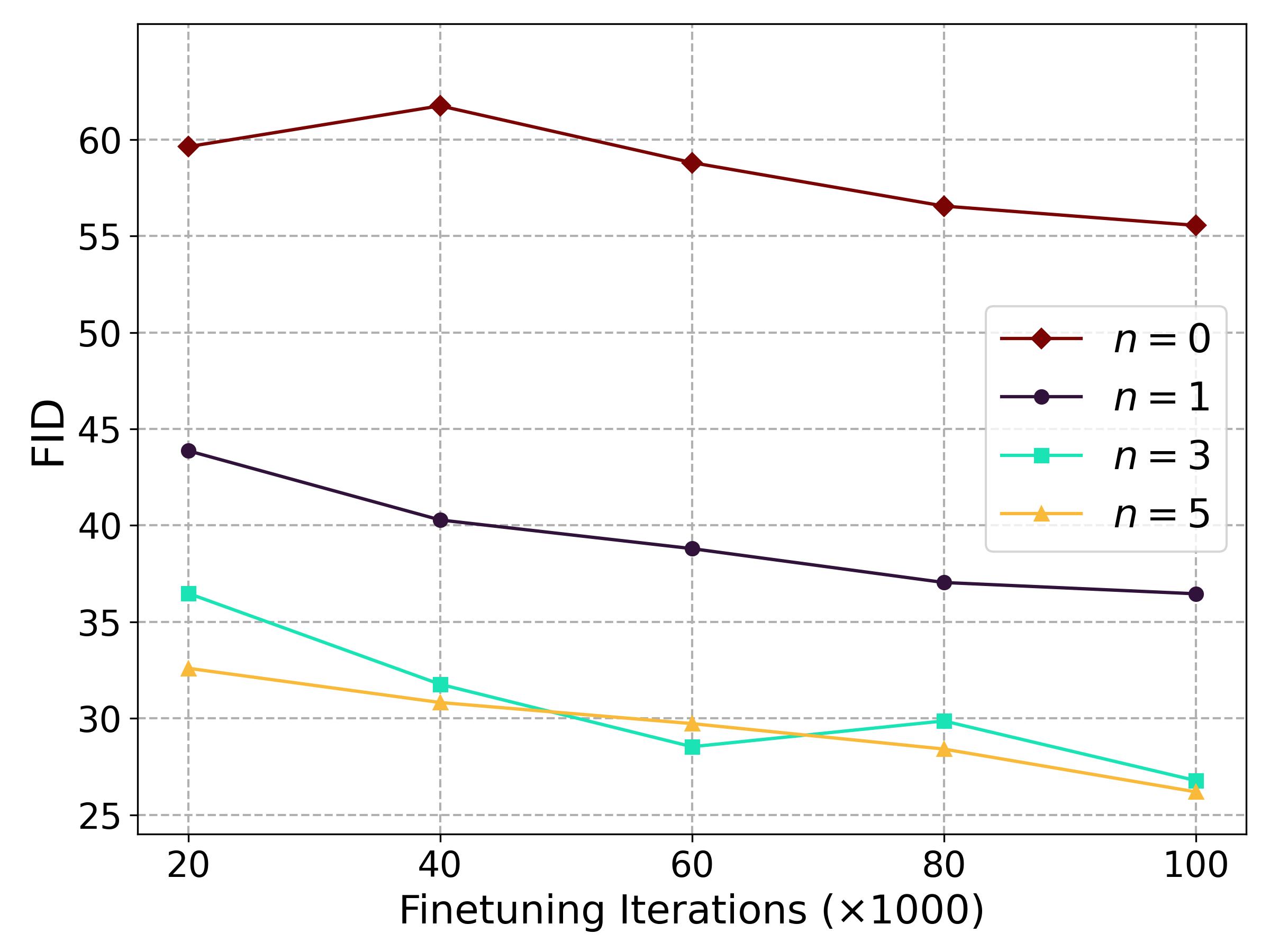}\tabularnewline
\end{tabular}
\par\end{centering}
\caption{Learning curves of finetuned dDR on Face-UMDT-Latent w.r.t. different
number of Tweedie refinement steps $n\in\left\{ 0,1,3,5\right\} $.
The task is Segment$\rightarrow$Sketch translation.\label{fig:loss_and_FID_curves_across_n}}
\end{figure}

\begin{figure}
\begin{centering}
\includegraphics[width=0.98\textwidth]{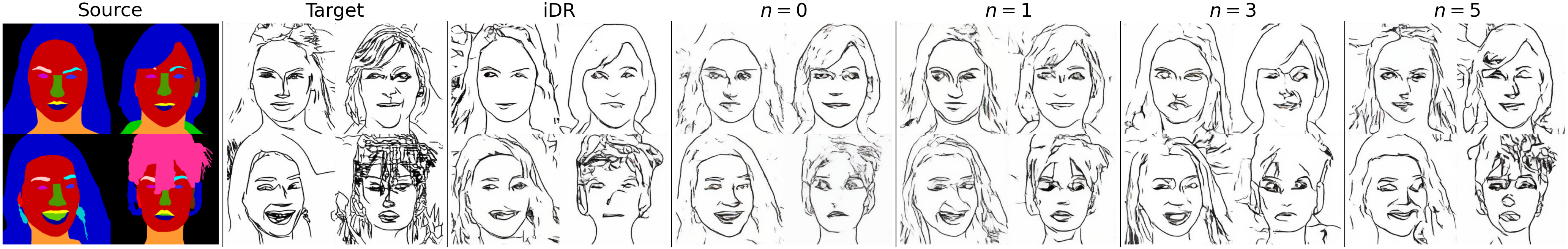}
\par\end{centering}
\caption{Comparison of translated images from Seg. to Ske. w.r.t different
value of $n$.\label{fig:abl_refine_steps}}
\end{figure}

We analyze the effect of the number of refinement steps \ensuremath{n}
n in the proposed Tweedie refinement (Eq.~\ref{eq:Tweedie_refinement})
by varying $n\in\left\{ 0,1,3,5\right\} $, with results summarized
in Table~\ref{tab:abl_refine_steps}. When $n=0$ (i.e., no refinement),
dDR performs poorly on direct translations between non-central domains
(e.g., Sketch$\leftrightarrow$Segment), though it remains comparable
to iDR on translations involving central domains (e.g., Sketch$\leftrightarrow$Face,
Segment$\leftrightarrow$Face). The failure is most pronounced for
the Segment$\rightarrow$Sketch translation task, where the target
domain has limited detail and the FID scores are highly sensitive
to minor variations. This indicates that without refinement, dDR struggles
to capture the correct mapping from Segment to Sketch. Increasing
$n$ consistently improves translation quality across all settings,
particularly in unpaired translations, as reflected in the performance
curves in Fig.~\ref{fig:loss_and_FID_curves_across_n} and the qualitative
results in Fig.~\ref{fig:abl_refine_steps}. The improvement arises
because the refined noisy sample $x_{t,(n)}^{j}$ provides a closer
approximation to samples from $p_{\text{ref}}\left(x_{t}^{j}|x^{c}\right)$,
thereby yielding more accurate predictions from $\epsilon_{\text{ref}}\left(x_{t}^{j},t,x^{c},j,c\right)$
(Eq.~\ref{eq:unpaired_loss}). Consequently, the unpaired loss $\Loss_{\text{unpaired}}$
better approximates the variational bound in Eq.~\ref{eq:VB_path_KL_div}.

\subsubsection{Impact of the coefficient $\lambda_{2}$ in the loss $\protect\Loss_{\text{final}}$\label{subsec:abla_impact_lambda}}

\begin{figure}
\begin{centering}
{\resizebox{\textwidth}{!}{%
\par\end{centering}
\begin{centering}
\begin{tabular}{ccc}
\includegraphics[width=0.33\textwidth]{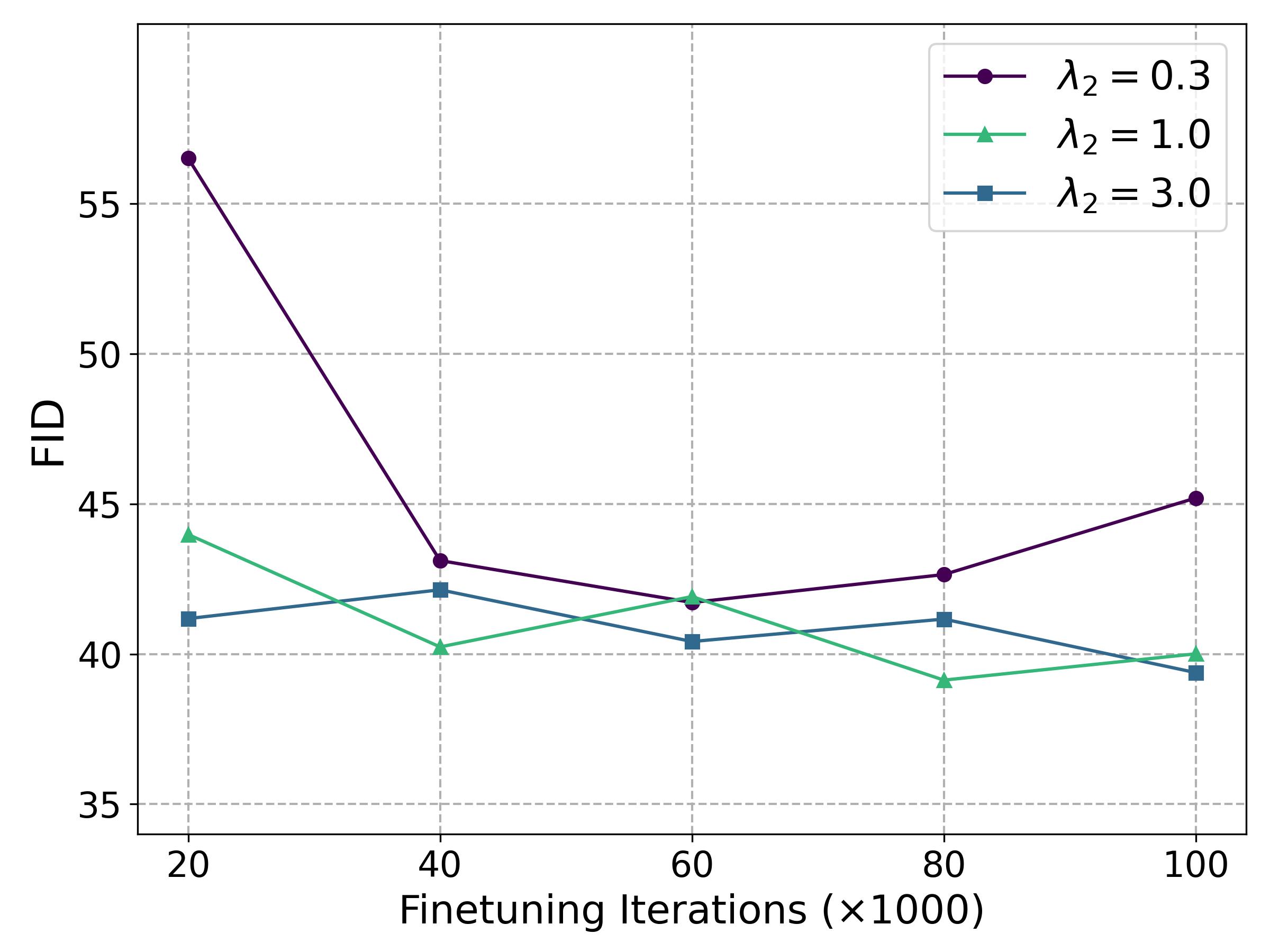} & \includegraphics[width=0.33\textwidth]{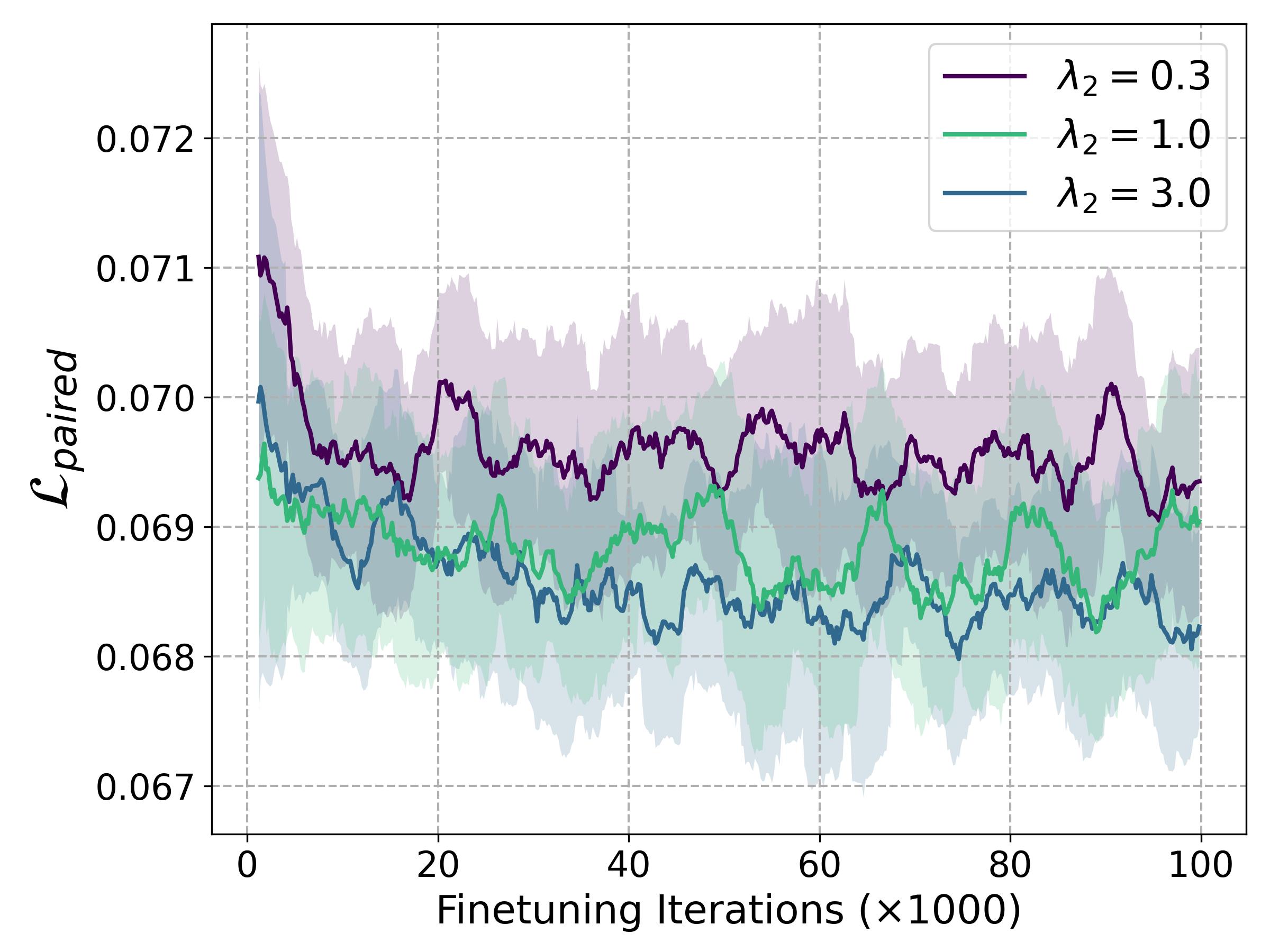} & \includegraphics[width=0.33\textwidth]{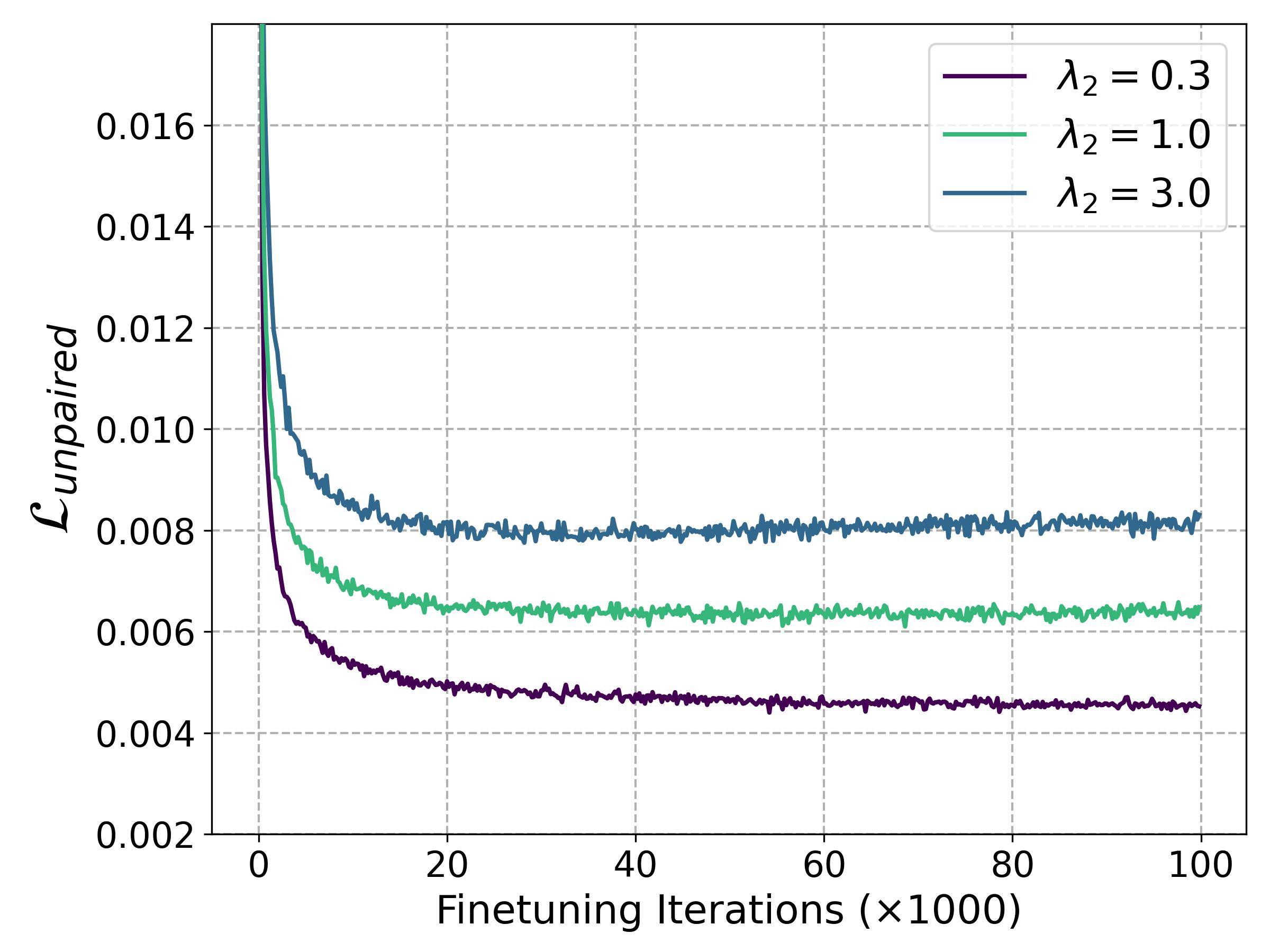}\tabularnewline
FID (Ske.$\to$Face) & $\mathcal{L}_{\text{paired}}$ on $t\in\left[0,250\right)$ (Ske.$\to$Face) & $\mathcal{L}_{\text{unpaired}}$ (Ske.$\to$Seg.)\tabularnewline
\end{tabular}}}
\par\end{centering}
\caption{Learning curves of dDR finetuned with different values of $\lambda_{2}$.\label{fig:Learning-curves_lambda2}}
\end{figure}

We investigate the effect of the coefficient $\lambda_{2}$ by experimenting
with different values in $\left\{ 0,0.3,1,3\right\} $. As shown in
Table~\ref{tab:ablation_lambda}, setting $\lambda_{2}=0$ causes
the FID scores for all translation tasks to diverge. This happens
because the unforgetting term $\mathcal{L}_{\text{paired}}$ is discarded
from $\mathcal{L}_{\text{final}}$, causing the finetuned dDR to forget
previously learned central$\leftrightarrow$non-central mappings (e.g.,
Sketch$\leftrightarrow$Face and Segment$\leftrightarrow$Face). Consequently,
non-central$\leftrightarrow$non-central translations such as Sketch$\leftrightarrow$Segment
are also learned incorrectly, since training them depends on noisy
samples from the central$\leftrightarrow$non-central mappings. When
$\lambda_{2}>0$, the fine-tuned model must balance preserving old
translations with learning new ones. Increasing $\lambda_{2}$ improves
the FID scores of the finetuned dDR on non-central$\leftrightarrow$non-central
translations, enabling them to match those of the pretrained iDR,
but at the cost of degrading performance on the central$\leftrightarrow$non-central
translations. This trade-off is more clearly reflected in the learning
curves in Fig.~\ref{fig:Learning-curves_lambda2}. Empirically, we
found that $\lambda_{2}=1$ provides the best balance across translation
tasks.

\subsubsection{Impact of training from scratch with $\protect\Loss_{\text{final}}$\label{subsec:Training-from-scratch}}

\begin{figure}
\begin{centering}
{\resizebox{\textwidth}{!}{%
\par\end{centering}
\begin{centering}
\begin{tabular}{ccc}
\includegraphics[width=0.33\textwidth]{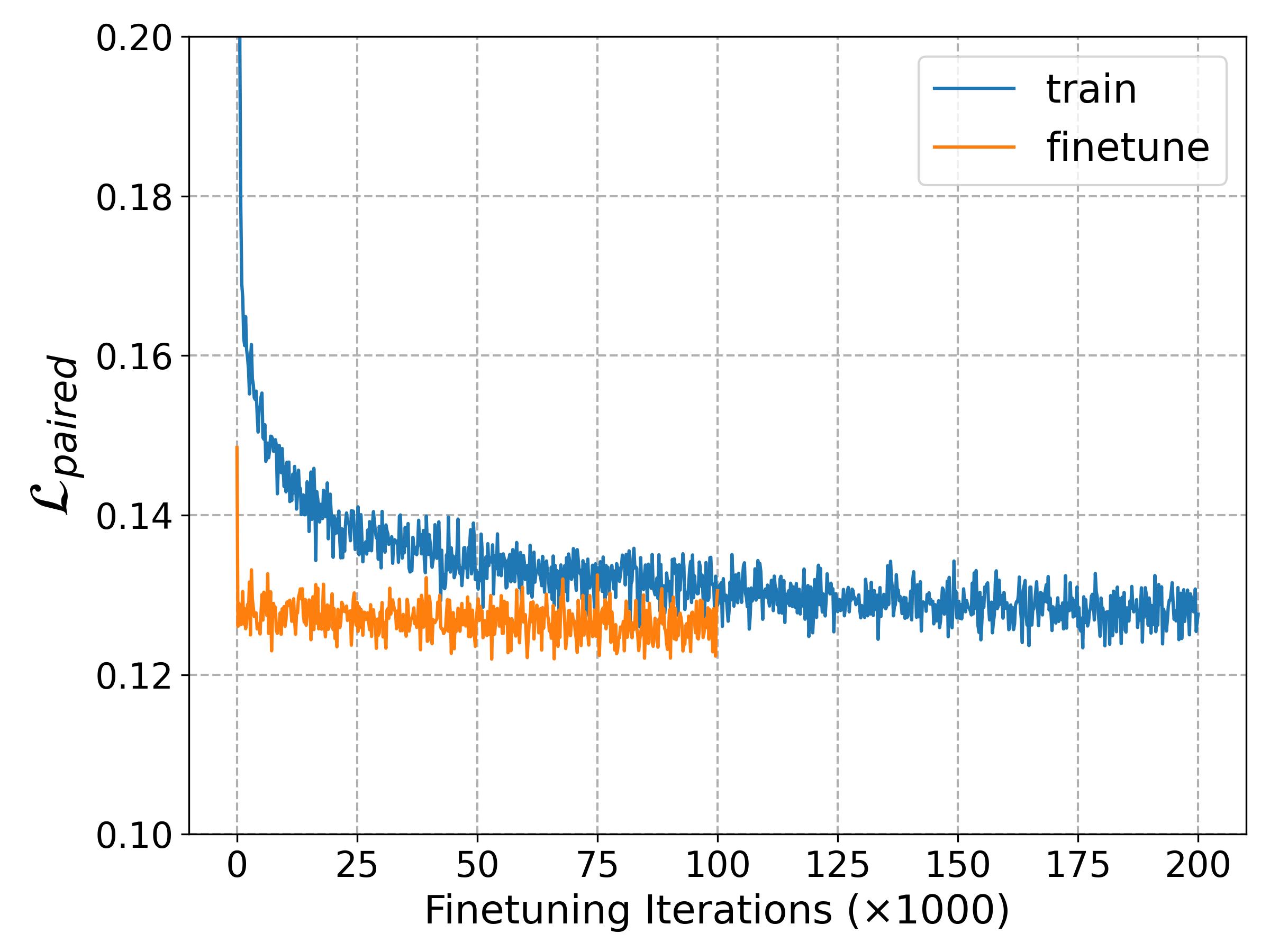} & \includegraphics[width=0.33\textwidth]{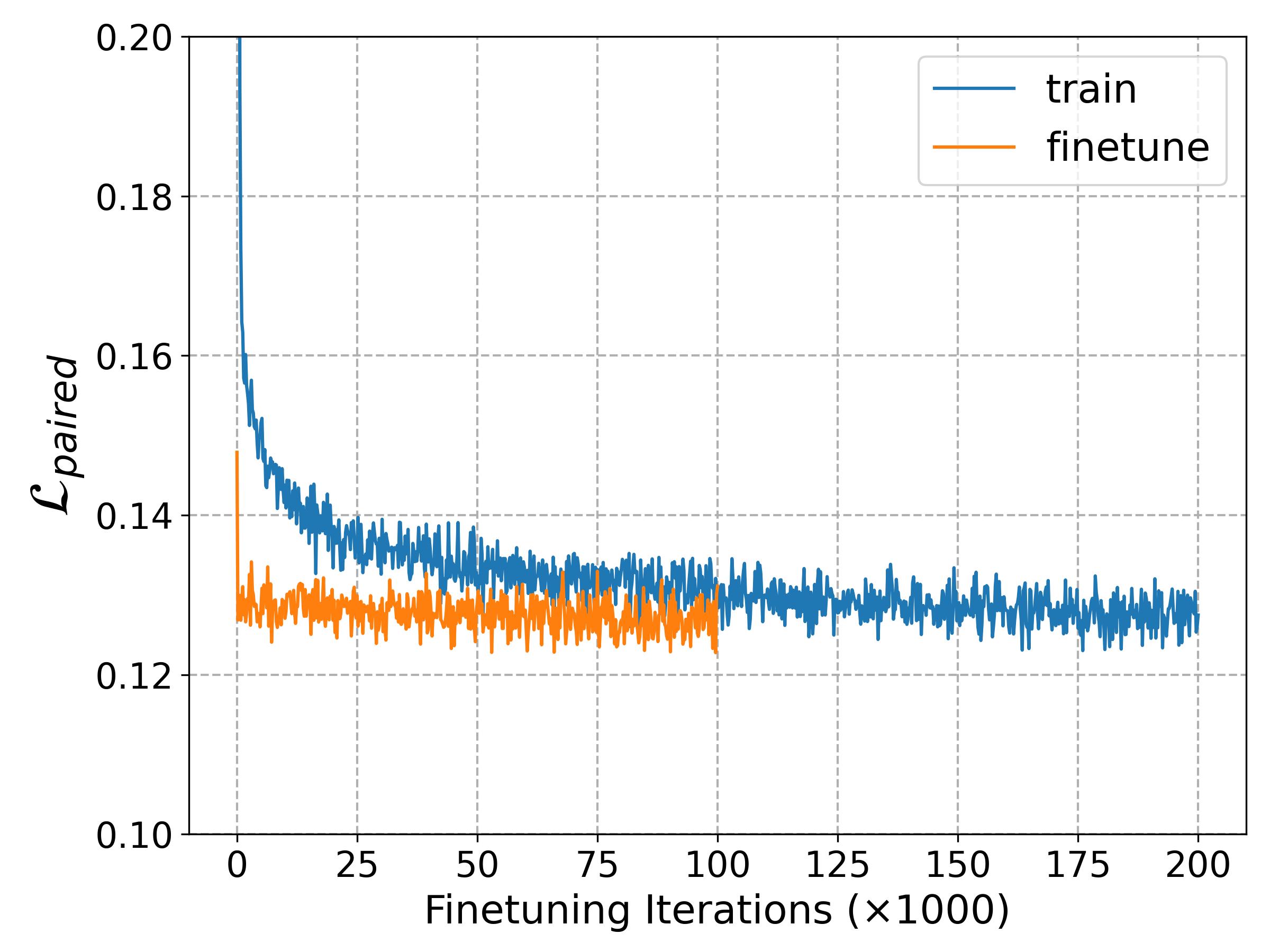} & \includegraphics[width=0.33\textwidth]{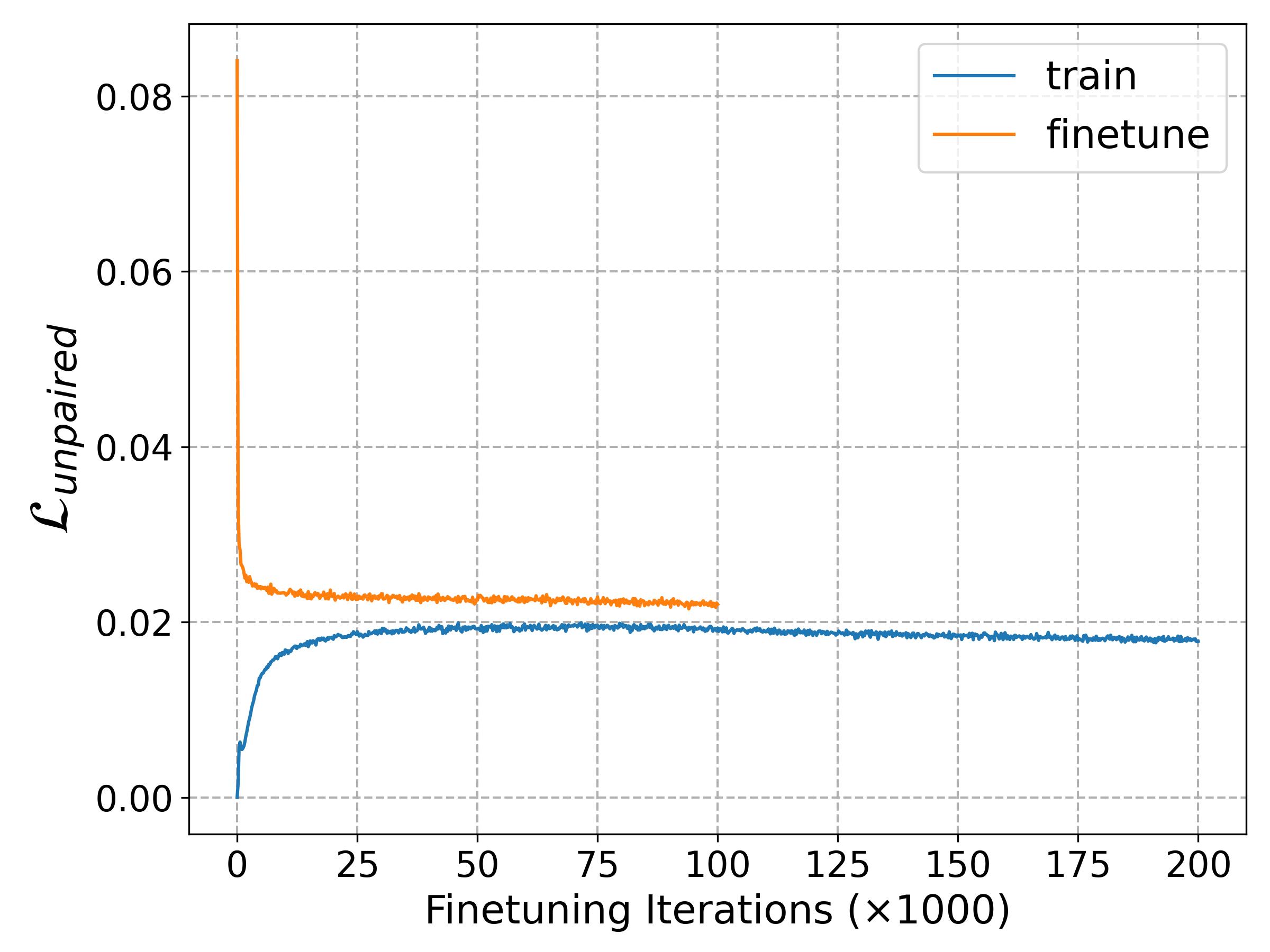}\tabularnewline
\includegraphics[width=0.33\textwidth]{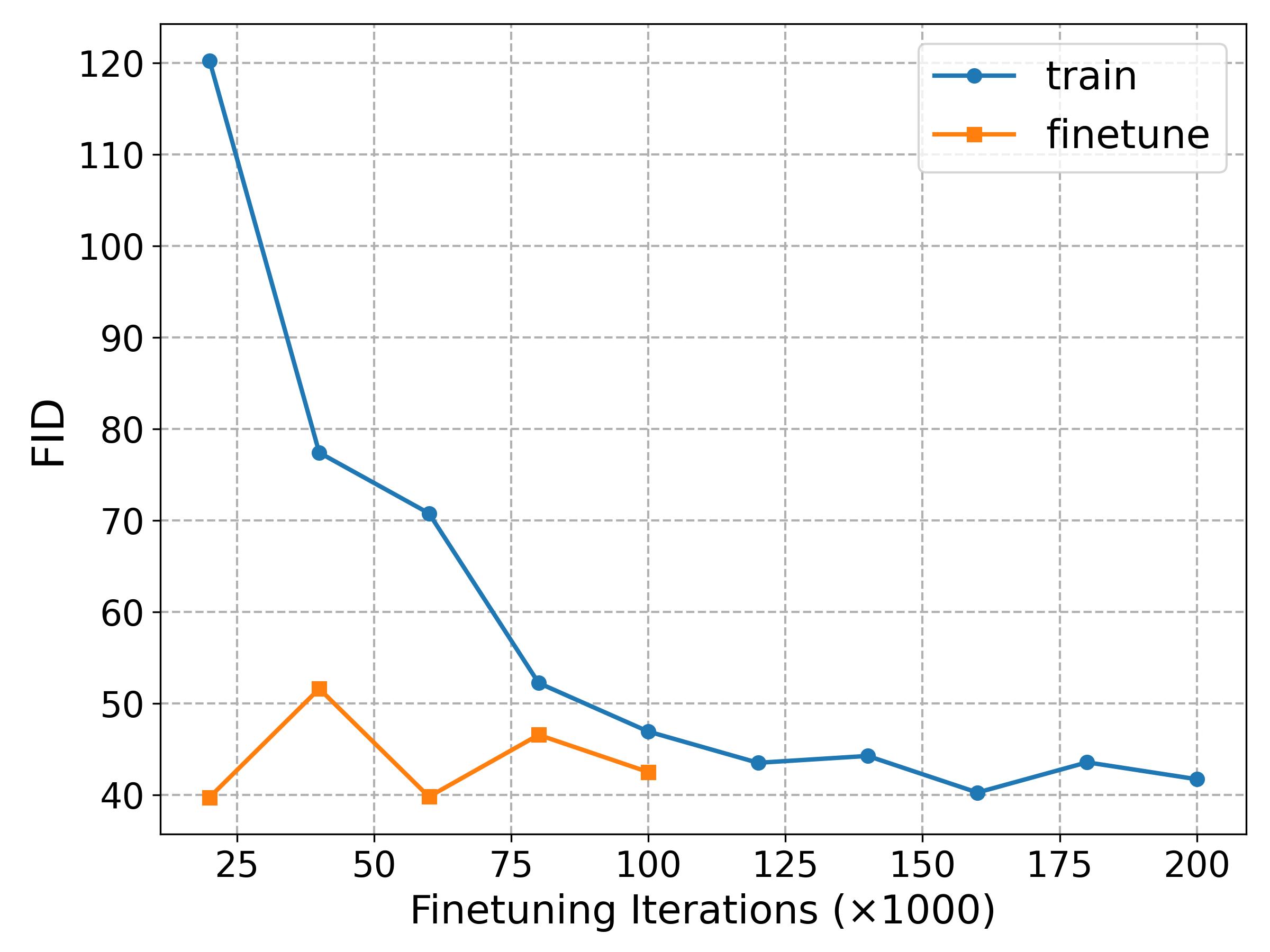} & \includegraphics[width=0.33\textwidth]{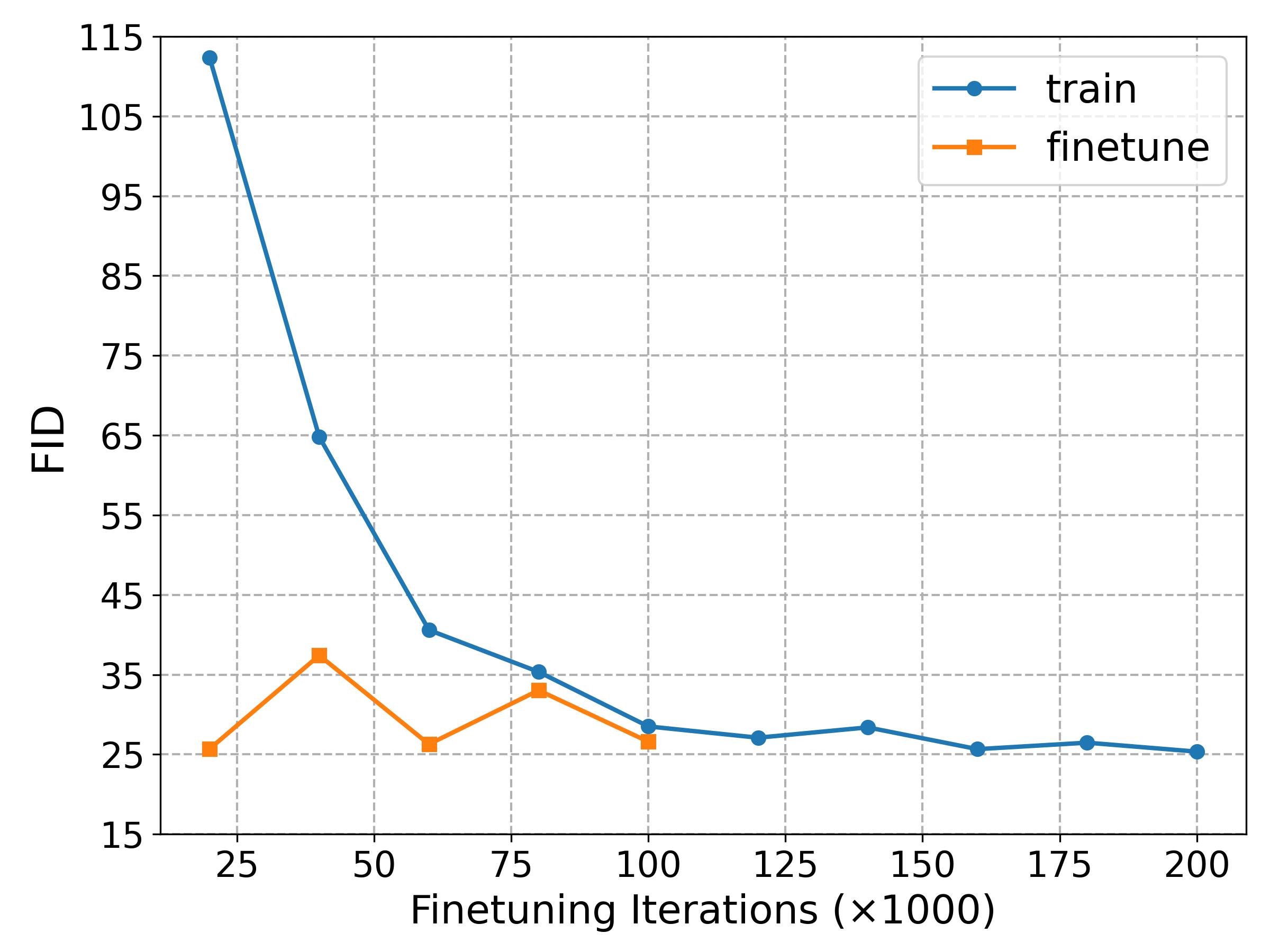} & \includegraphics[width=0.33\textwidth]{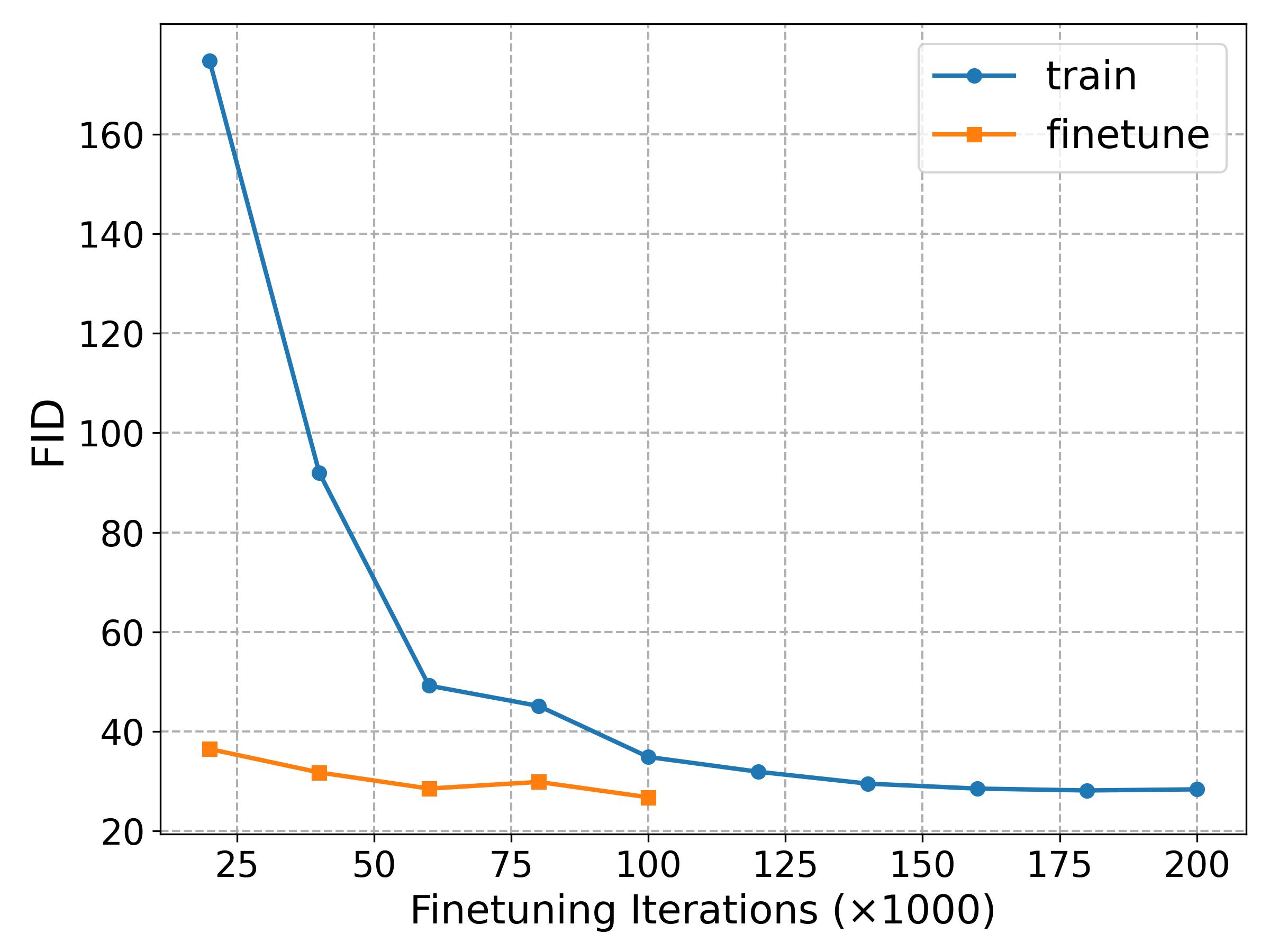}\tabularnewline
(a) Sketch$\to$Face & (b) Segment$\to$Face & (c) Segment$\to$Sketch\tabularnewline
\end{tabular}}}
\par\end{centering}
\caption{Loss and FID curves of trained-from-scratch and finetuned dDRs.\label{fig:curves_dDR_trained_and_finetuned}}
\end{figure}

We study the effect of training dDR from scratch versus fine-tuning
it from a pretrained iDR, both strategies using roughly the same loss
$\Loss_{\text{final}}$ in Eq.~\ref{eq:final_loss}. In this experiment,
we set $n=3$. Training from scratch is computationally demanding--requiring
over three days to run 200K iterations of a small U-Net on a single
H100 GPU--so we limit the from-scratch run to 200K iterations and
compare it with 100K fine-tuning iterations under identical settings.
Fig.~\ref{fig:curves_dDR_trained_and_finetuned} presents the loss
and FID curves for both strategies on central$\leftrightarrow$non-central
and non-central$\leftrightarrow$non-central translation tasks. Overall,
dDR trained from scratch performs poorly in the early stages (as indicated
by high FID scores), but gradually improves and eventually approaches
the performance of the fine-tuned variant as training converges. Interestingly,
the $\mathcal{L}_{\text{unpaired}}$ curve for from-scratch training
behaves differently from $\mathcal{L}_{\text{paired}}$: it increases
from near zero rather than decreasing from a large value. This occurs
because, at the start of training, $\epsilon_{\theta}$ is untrained,
making the expectations of $\epsilon_{\theta}\left(x_{t}^{j},t,x^{i},j,i\right)$
and of $\epsilon_{\theta}\left(x_{t}^{j},t,x^{c},j,c\right)$ approximately
equal.

\subsubsection{Comparison of iDR and dDR across sampling steps\label{subsec:abla_different_NFE}}

\begin{table*} \centering \resizebox{\textwidth}{!}{ \begin{tabular}{c|cc|cc|cc|cc} \toprule \multirow{3}{*}{NFE} & \multicolumn{2}{c|}{\textbf{Faces-UMDT-Latent}} & \multicolumn{6}{c}{\textbf{COCO-UMDT-Star}} \\ \cmidrule(lr){2-3}\cmidrule(lr){4-9}  & \multicolumn{2}{c|}{Ske.\(\leftrightarrow\)Seg.} & \multicolumn{2}{c|}{Ske.\(\leftrightarrow\)Seg.} & \multicolumn{2}{c|}{Ske.\(\leftrightarrow\)Depth} & \multicolumn{2}{c}{Seg.\(\leftrightarrow\)Depth} \\ \cmidrule(lr){2-3}\cmidrule(lr){4-5}\cmidrule(lr){6-7}\cmidrule(lr){8-9}  & iDR & dDR & iDR & dDR & iDR & dDR & iDR & dDR \\ \midrule 10   & 74.33/7.95 & 39.29/6.65 & 98.14/79.17 & 31.29/34.19 & 78.65/13.96 & 27.10/12.09 & 79.95/24.97 & 35.31/16.48 \\ 20   & 42.25/6.98 & 23.79/5.82 & 50.09/32.07 & 28.26/25.67 & 42.95/11.30 & 22.69/11.26 & 32.93/15.35 & 26.03/15.63 \\ 30   & 31.77/6.67 & 22.31/5.72 & 36.01/25.99 & 27.93/24.81 & 31.37/9.83 & 22.09/10.35 & 26.12/14.71 & 25.57/15.02 \\ 50   & 23.37/6.39 & 21.85/5.61 & 27.84/23.45 & 27.19/24.06 & 19.39/9.03 & 21.65/9.59 & 24.87/13.93 & 25.21/14.74 \\ 100  & 18.05/6.25 & 20.82/5.58 & 24.70/23.16 & 26.87/23.78 & 18.81/8.79 & 21.18/9.29  & 24.08/12.49 & 25.18/14.62 \\ 1000 & 16.17/6.19 & 19.42/5.52 & 23.06/23.01 & 26.73/23.64 & 18.15/8.86 & 20.75/9.42  & 23.37/12.11 & 24.91/14.87 \\ \bottomrule \end{tabular}} \caption{FID across NFE for non-central\(\leftrightarrow\)non-central translations with iDR and dDR on Faces-UMDT-Latent and COCO-UMDT-Star.} \label{tab:ablation_NFE} \end{table*} 

We evaluate iDR and dDR on Faces-UMDT-Latent and COCO-UMDT-Star while
varying the total number of sampling steps (NFE) from 10 to 1000.
We focus on non-central$\leftrightarrow$non-central translations
because the central$\leftrightarrow$non-central results for iDR and
dDR are already comparable at matched NFE in the main experiments.
Both methods use 304.8M-parameter U-Net models and are evaluated with
the same total NFE for each non-central$\leftrightarrow$non-central
translation.

From Table \ref{tab:ablation_NFE}, we observe that iDR\textquoteright s
performance deteriorates sharply as NFE decreases, whereas dDR remains
much more stable, with only a moderate drop in FID. At NFE \ensuremath{\le}
30, dDR achieves 50--100\% lower FID scores than iDR on several translation
tasks, such as Seg.$\rightarrow$Ske. on Faces-UMDT-Latent, and Ske.$\leftrightarrow$Seg.,
Ske.$\leftarrow$Depth, and Seg.$\leftarrow$Depth on COCO-UMDT-Star.
This discrepancy arises because iDR relies on an intermediate central-domain
sample whose quality degrades significantly when NFE is small, thereby
impairing the final non-central-domain output. By contrast, dDR directly
generates the non-central domain and thus avoids this issue. These
results strongly indicate that dDR is not sensitive to the central
domain sample quality and is the clearly preferable choice when sampling
with a limited number of steps.

\subsubsection{Comparison of Standard and Bridge-Based DRs\label{subsec:Comparison-of-Standard}}

\begin{table*}
\begin{centering}
{\resizebox{\textwidth}{!}{%
\par\end{centering}
\begin{centering}
\begin{tabular}{cccccccc}
\toprule 
\multirow{2}{*}{Model type} & \multirow{2}{*}{Model} & \multicolumn{3}{c}{FID$\downarrow$} & \multicolumn{3}{c}{LPIPS$\downarrow$}\tabularnewline
\cmidrule{3-8} \cmidrule{4-8} \cmidrule{5-8} \cmidrule{6-8} \cmidrule{7-8} \cmidrule{8-8} 
 &  & Ske.$\leftrightarrow$Face & Seg.$\leftrightarrow$Face & \textcolor{brown}{Ske.$\leftrightarrow$Seg.} & Ske.$\leftrightarrow$Face & Seg.$\leftrightarrow$Face & \textcolor{brown}{Ske.$\leftrightarrow$Seg.}\tabularnewline
\midrule
\midrule 
\multirow{4}{*}{Bridge} & BBDM & 25.69/41.79 & 19.02/27.76 & \textcolor{brown}{27.92/20.76} & 0.244/0.457 & 0.144/0.492 & \textcolor{brown}{0.399/0.196}\tabularnewline
 & I$^{2}$SB & 16.56/31.81 & 10.58/24.56 & \textcolor{brown}{21.37/13.16} & 0.241/0.456 & 0.153/0.492 & \textcolor{brown}{0.396/0.208}\tabularnewline
 & DDBM & 15.51/42.55 & 10.32/27.77 & \textcolor{brown}{19.51/11.41} & 0.246/0.477 & 0.156/0.506 & \textcolor{brown}{0.408/0.213}\tabularnewline
 & BDBM & 13.67/33.98 & 6.12/26.73 & \textcolor{brown}{24.88/6.63} & 0.244/0.479 & 0.151/0.500 & \textcolor{brown}{0.382/0.204}\tabularnewline
\midrule 
Diffusion & DDPM & \textbf{9.07}/\textbf{23.88} & \textbf{6.12}/\textbf{19.12} & \textbf{\textcolor{brown}{15.37}}\textcolor{brown}{/}\textbf{\textcolor{brown}{6.15}} & \textbf{0.221}/\textbf{0.427} & \textbf{0.129}/\textbf{0.471} & \textbf{\textcolor{brown}{0.377}}\textcolor{brown}{/}\textbf{\textcolor{brown}{0.177}}\tabularnewline
\bottomrule
\end{tabular}}}
\par\end{centering}
\caption{Results on Faces-UMDT-Latent comparing diffusion-based and bridge-based
DRs.\label{tab:abl_diff_vs_bridge}}
\end{table*}

\begin{figure}
\begin{centering}
\includegraphics[width=0.95\textwidth]{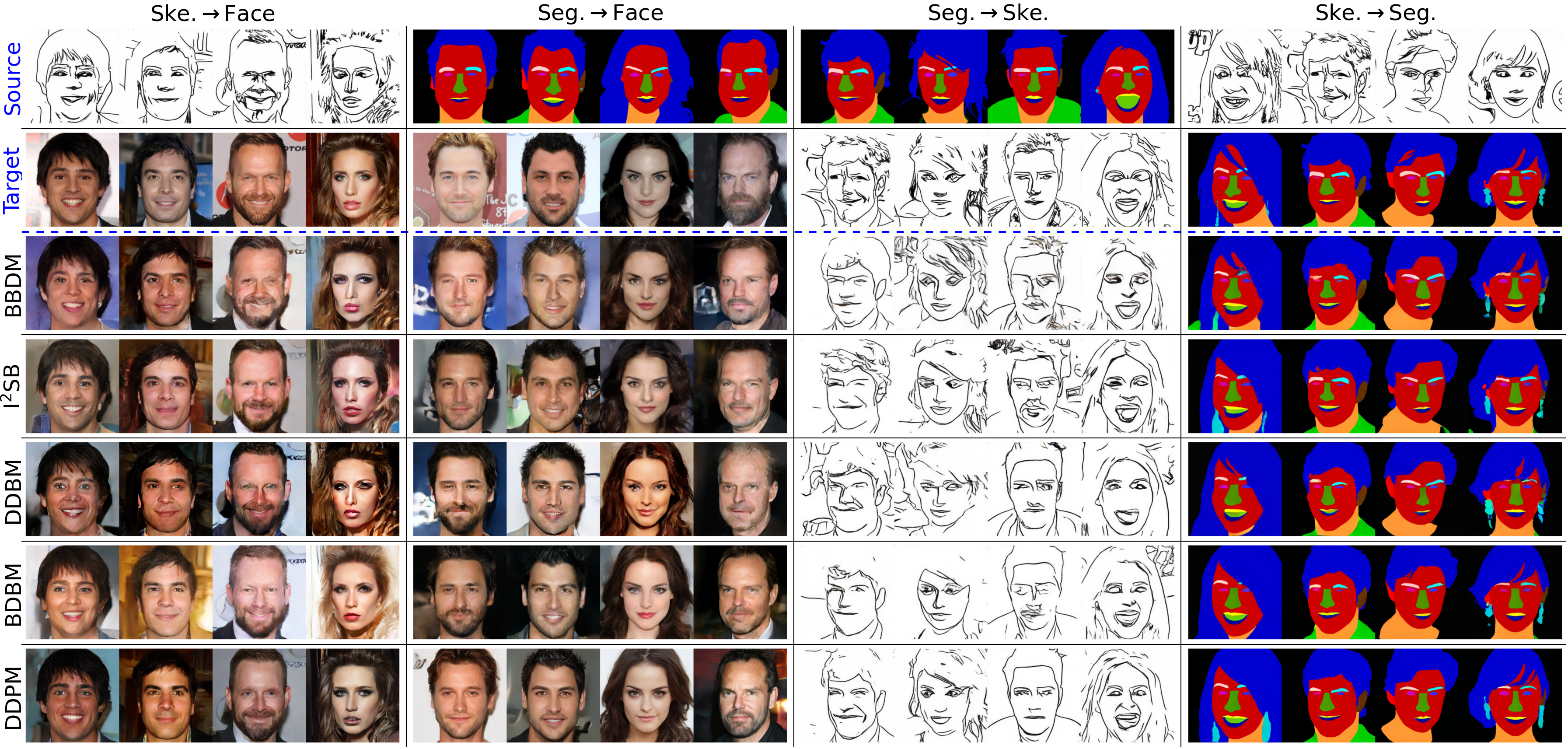}
\par\end{centering}
\caption{Qualitative comparison of diffusion-based and bridge-based DRs on
Faces-UMDT-Latent.\label{fig:visualization_bridge_vs_diff}}
\end{figure}

We compare diffusion-based against bridge-based DRs on Faces-UMDT-Latent.
For diffusion-based DR, we adopt DDPM schedules with the noise parameterization.
For bridge-based DR, we follow schedules and parameterizations of
state-of-the-art bridges, including BBDM \cite{LiX0L23}, I$^{2}$SB
\cite{LiuVHTNA23}, DDBM \cite{zhou2024denoising}, and BDBM \cite{Kieu2025}.
In all cases, a single 304.8M-parameter U-Net models all conditionals
between the central and each non-central domain; translations between
two non-central domains are executed indirectly via the central domain.

Table \ref{tab:abl_diff_vs_bridge} shows that the diffusion-based
DR outperforms bridge variants on most translation directions, with
especially large margins on high-variance mappings (e.g., Ske.$\to$Face,
Seg.$\to$Face). We attribute this advantage to diffusion\textquoteright s
Gaussian prior and iterative denoising, which recover missing detail
under sparse conditioning (e.g., sketches), whereas bridge-based sampling
tightly anchors to the input and can suppresses plausible completions.
Consistent with this, Fig.~\ref{fig:visualization_bridge_vs_diff}
shows that bridge-based DRs on Ske.$\to$Face often over-condition
on noisy sketch details, producing unrealistic artifacts. Moreover,
diffusion models partially share generative processes across domains
(from a common Gaussian), while bridge models learn separate source-to-target
couplings for each paired dataset. A single U-Net thus faces a harder
multi-coupling approximation under bridge designs, particularly when
paired datasets differ substantially. Overall, these results indicate
that diffusion-based DRs are better suited to MDT and can be fine-tuned
to support direct non-central translations without paired supervision.

\subsubsection{Impact of the learning rate for finetuning}

Table \ref{tab:ablation_lr} examines the learning-rate trade-off
when finetuning iDR into dDR for direct translation. Higher rates
accelerate learning of the new Ske.$\leftrightarrow$Seg. mappings
but significantly degrade the pretrained mappings, e.g., Ske.$\leftrightarrow$Face
and Seg.$\leftrightarrow$Face, relative to iDR. At 1e-4, dDR achieves
the best FID on Ske.$\leftrightarrow$Seg. yet yields the worst FIDs
on Ske.$\leftrightarrow$Face and Seg.$\leftrightarrow$Face compared
with iDR and dDR finetuned at lower rates. Reducing the rate reduces
catastrophic forgetting and restores the pretrained directions, reaching
14.90/39.94 with Ske.$\leftrightarrow$Face and 13.01/25.12 on Seg.$\leftrightarrow$Face
at 1e-6. This improvement comes at the cost of the direct translation
mappings, with Ske.$\to$Seg. FID increasing from 52.41 to 82.04 and
Seg.$\to$Ske. FID increasing from 12.87 to 18.84. Overall, the results
show a clear trade-off between retaining pretrained tasks and learning
the new one. Among the tested settings, 5e-5 offers the best balance.
\begin{center}
\begin{table}
\begin{centering}
{\resizebox{0.48\textwidth}{!}{%
\par\end{centering}
\begin{centering}
\begin{tabular}{ccccc}
\toprule 
\multirow{2}{*}{} & \multirow{2}{*}{lr} & \multicolumn{3}{c}{FID$\downarrow$}\tabularnewline
\cmidrule{3-5} \cmidrule{4-5} \cmidrule{5-5} 
 &  & Ske.$\leftrightarrow$Face & Seg.$\leftrightarrow$Face & \textcolor{brown}{Ske.$\leftrightarrow$Seg.}\tabularnewline
\midrule
\midrule 
iDR & - & 14.21/39.06 & 10.18/24.95 & \textcolor{brown}{20.82/10.85}\tabularnewline
\midrule 
\multirow{5}{*}{dDR} & 1e-4 & 17.05/43.18 & 15.16/28.90 & \textcolor{brown}{52.41/12.87}\tabularnewline
\cmidrule{2-5} \cmidrule{3-5} \cmidrule{4-5} \cmidrule{5-5} 
 & 5e-5 & 16.11/39.82 & 13.75/25.77 & \textcolor{brown}{55.30/13.35}\tabularnewline
\cmidrule{2-5} \cmidrule{3-5} \cmidrule{4-5} \cmidrule{5-5} 
 & 1e-5 & 15.55/39.56 & 13.60/25.13 & \textcolor{brown}{71.09/15.09}\tabularnewline
\cmidrule{2-5} \cmidrule{3-5} \cmidrule{4-5} \cmidrule{5-5} 
 & 5e-6 & 14.58/40.01 & 13.51/24.58 & \textcolor{brown}{76.96/16.50}\tabularnewline
\cmidrule{2-5} \cmidrule{3-5} \cmidrule{4-5} \cmidrule{5-5} 
 & 1e-6 & 14.90/39.94 & 13.01/25.12 & \textcolor{brown}{82.04/18.84}\tabularnewline
\bottomrule
\end{tabular}}}
\par\end{centering}
\caption{FID on Faces-UMDT-Latent for dDR across finetuning learning rates.
All dDR models were fine-tuned without Tweedie refinement ($n=0$).
\label{tab:ablation_lr}}
\end{table}
\par\end{center}

\subsection{Qualitative results on COCO-UMDT-Star and COCO-UMDT-Chain}

\begin{figure}
\begin{centering}
\subfloat[Results on COCO-UMDT-Star. The classes for each segmentation map (from
top to bottom) are: (\textcolor{red}{snow}, \textcolor{green}{person}),
(\textcolor{red}{wall}, \textcolor{green}{person}, \textcolor{blue}{floor}),
(\textcolor{red}{cow}, \textcolor{blue}{grass}).]{\begin{centering}
\includegraphics[width=0.95\textwidth]{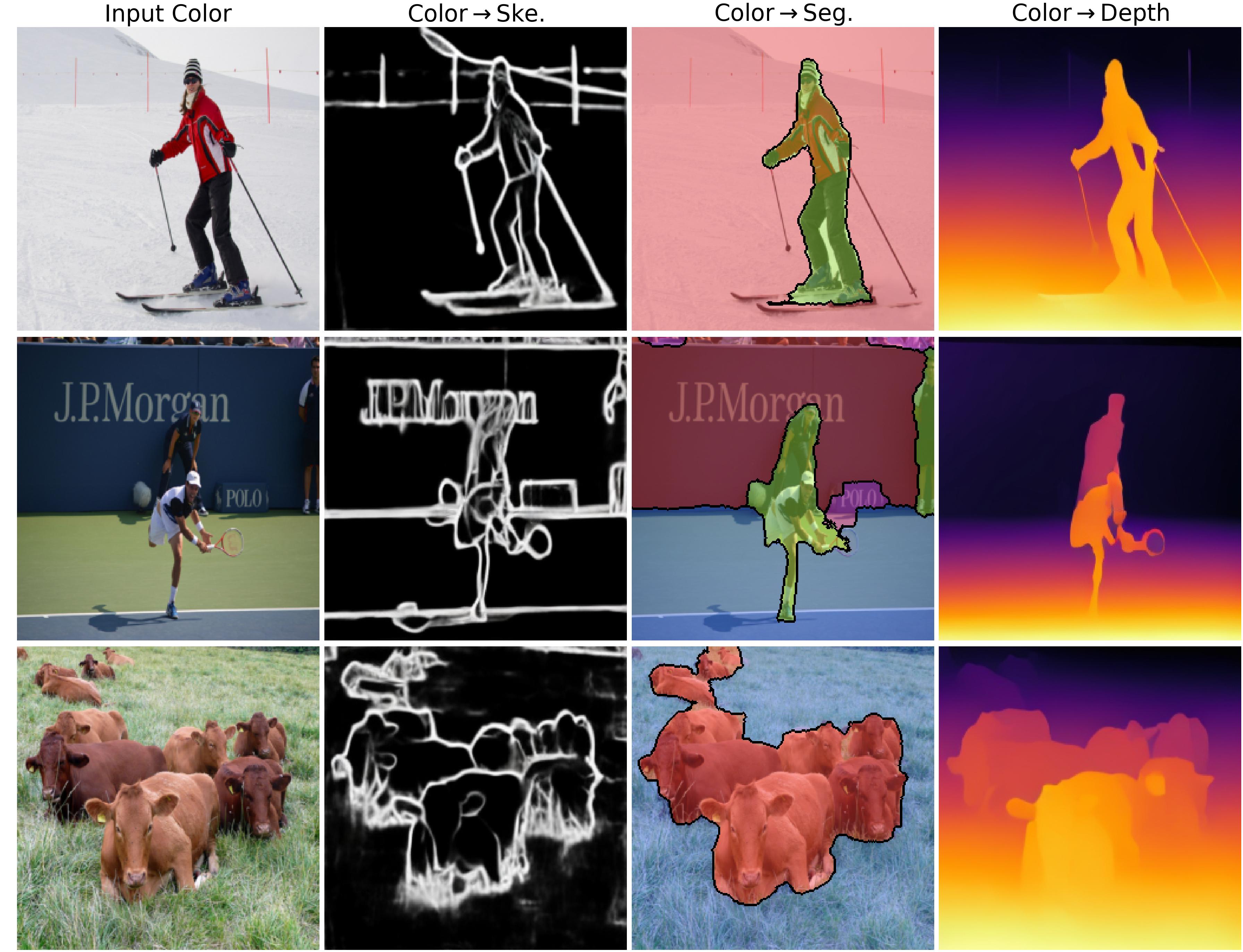}
\par\end{centering}
}
\par\end{centering}
\begin{centering}
\subfloat[Results on COCO-UMDT-Chain. The classes for each segmentation map
(from top to bottom) are: (\textcolor{red}{snow}, \textcolor{green}{person},
\textcolor{blue}{sky}), (\textcolor{red}{wall}, \textcolor{green}{person},
\textcolor{blue}{floor}), (\textcolor{red}{cow}, \textcolor{blue}{grass}).]{\begin{centering}
\includegraphics[width=0.95\textwidth]{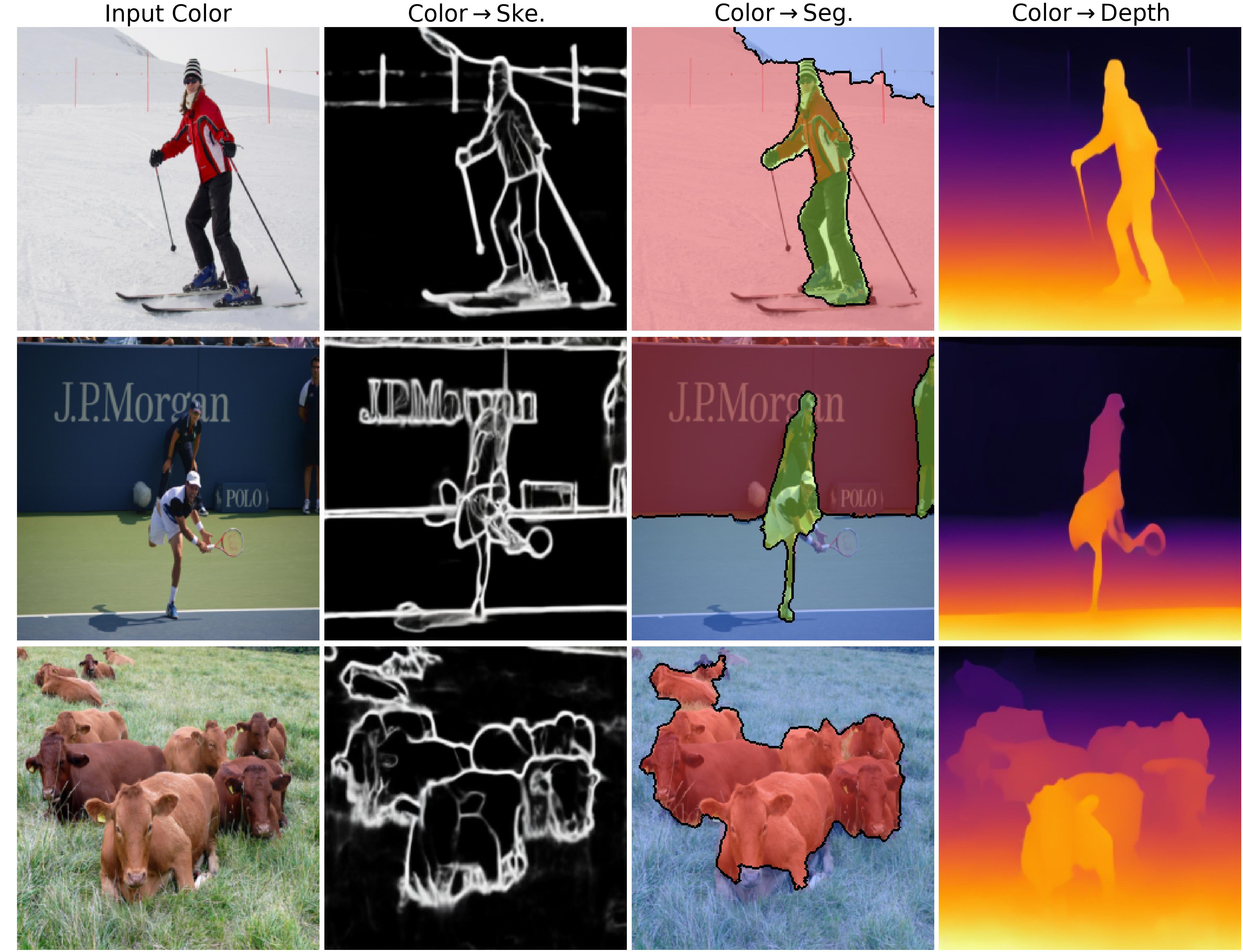}
\par\end{centering}
}
\par\end{centering}
\caption{Results of finetuned dDR on COCO-UMDT-Star (a) and COCO-UMDT-Chain
(b) for translation tasks from Color domain to Sketch, Segment, and
Depth domains.\label{fig:dDR_coco_color}}

\end{figure}

\begin{figure}
\begin{centering}
\subfloat[Results on COCO-UMDT-Star. The classes for each segmentation map (from
top to bottom) are: (\textcolor{magenta}{building}\textcolor{black}{,}\textcolor{red}{{}
}\textcolor{orange}{bus}, \textcolor{blue}{road}, \textcolor{red}{bush}),
(\textcolor{green}{tree}, \textcolor{blue}{train}), (\textcolor{green}{elephant},
\textcolor{red}{sky}, \textcolor{blue}{gravel}).]{\begin{centering}
\includegraphics[width=0.95\textwidth]{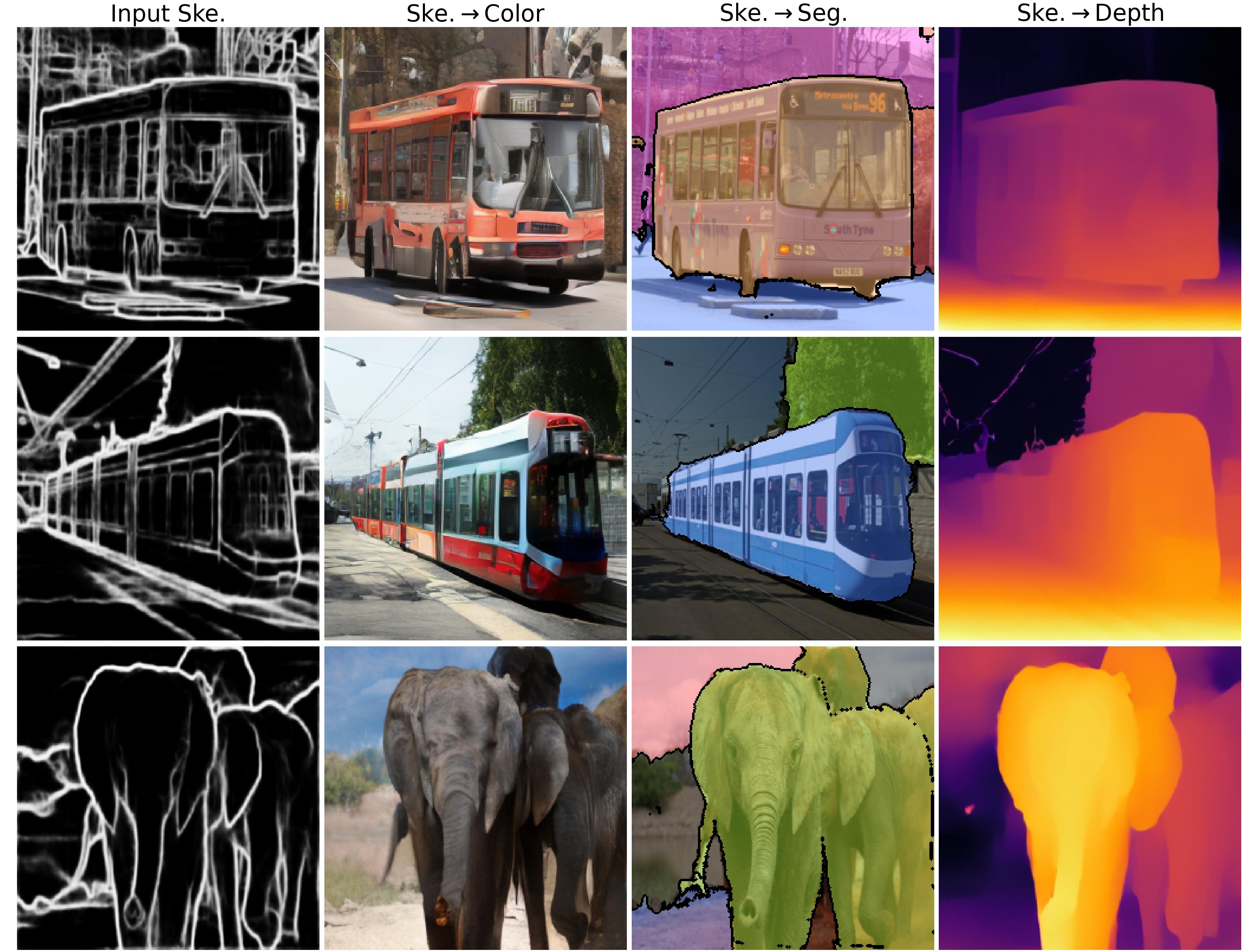}
\par\end{centering}
}
\par\end{centering}
\begin{centering}
\subfloat[Results on COCO-UMDT-Chain. The classes for each segmentation map
(from top to bottom) are: (\textcolor{green}{bus}, \textcolor{blue}{road},
\textcolor{red}{wall}), (\textcolor{green}{tree}, \textcolor{blue}{train},
\textcolor{magenta}{sky}), (\textcolor{green}{elephant}, \textcolor{red}{sky},
\textcolor{blue}{tree}).]{\begin{centering}
\includegraphics[width=0.95\textwidth]{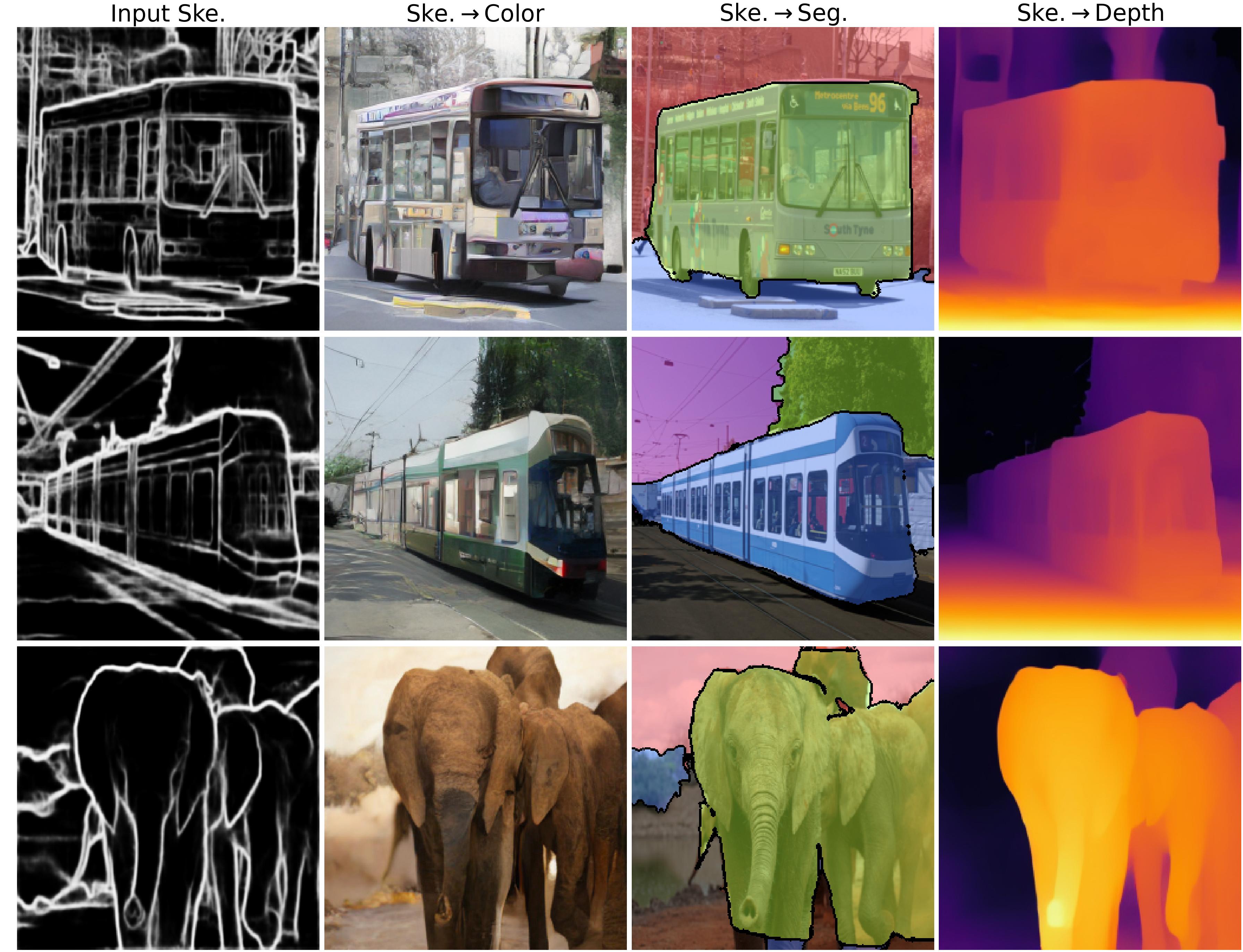}
\par\end{centering}

}
\par\end{centering}
\begin{centering}
\caption{Results of finetuned dDR on COCO-UMDT-Star (a) and COCO-UMDT-Chain
(b) for translation tasks from Sketch domain to Color, Segment, and
Depth domains.\label{fig:dDR_coco_ske}}
\par\end{centering}
\end{figure}

\begin{figure}
\begin{centering}
\subfloat[Results on COCO-UMDT-Star. The classes for each segmentation map (from
top to bottom) are: (\textcolor{blue}{sky}\textcolor{black}{, }\textcolor{green}{airplane}\textcolor{black}{,}\textcolor{red}{{}
bridge}), (\textcolor{orange}{airplane}), (\textcolor{orange}{banana},
\textcolor{blue}{table}).]{\begin{centering}
\includegraphics[width=0.95\textwidth]{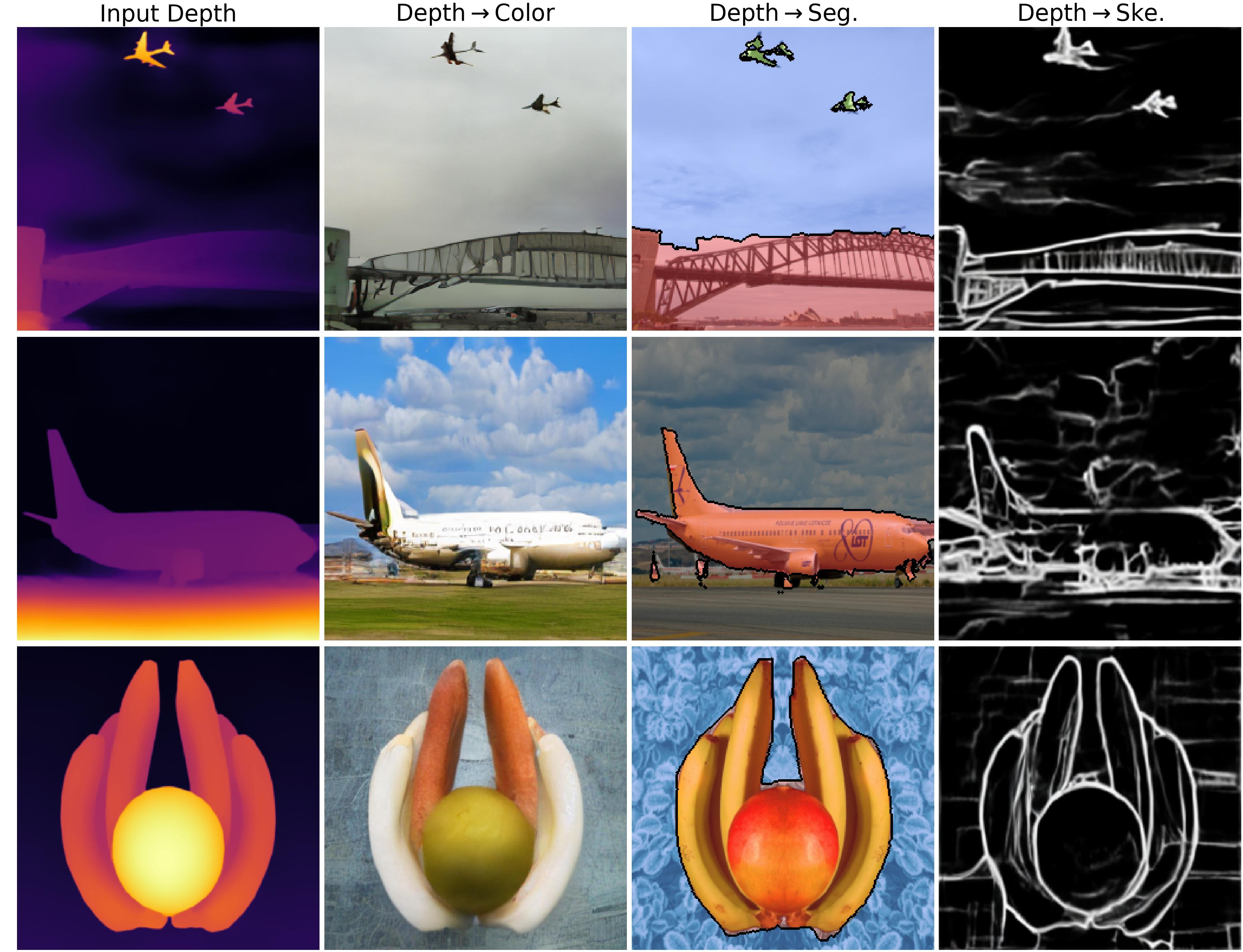}
\par\end{centering}
}
\par\end{centering}
\begin{centering}
\subfloat[Results on COCO-UMDT-Chain. The classes for each segmentation map
(from top to bottom) are: (\textcolor{red}{bridge}, \textcolor{green}{airplane},
\textcolor{blue}{sky}, \textcolor{magenta}{undefined}), (\textcolor{orange}{airplane},
\textcolor{green}{sky}, \textcolor{blue}{road}), (\textcolor{orange}{banana},
\textcolor{blue}{floor}).]{\begin{centering}
\includegraphics[width=0.95\textwidth]{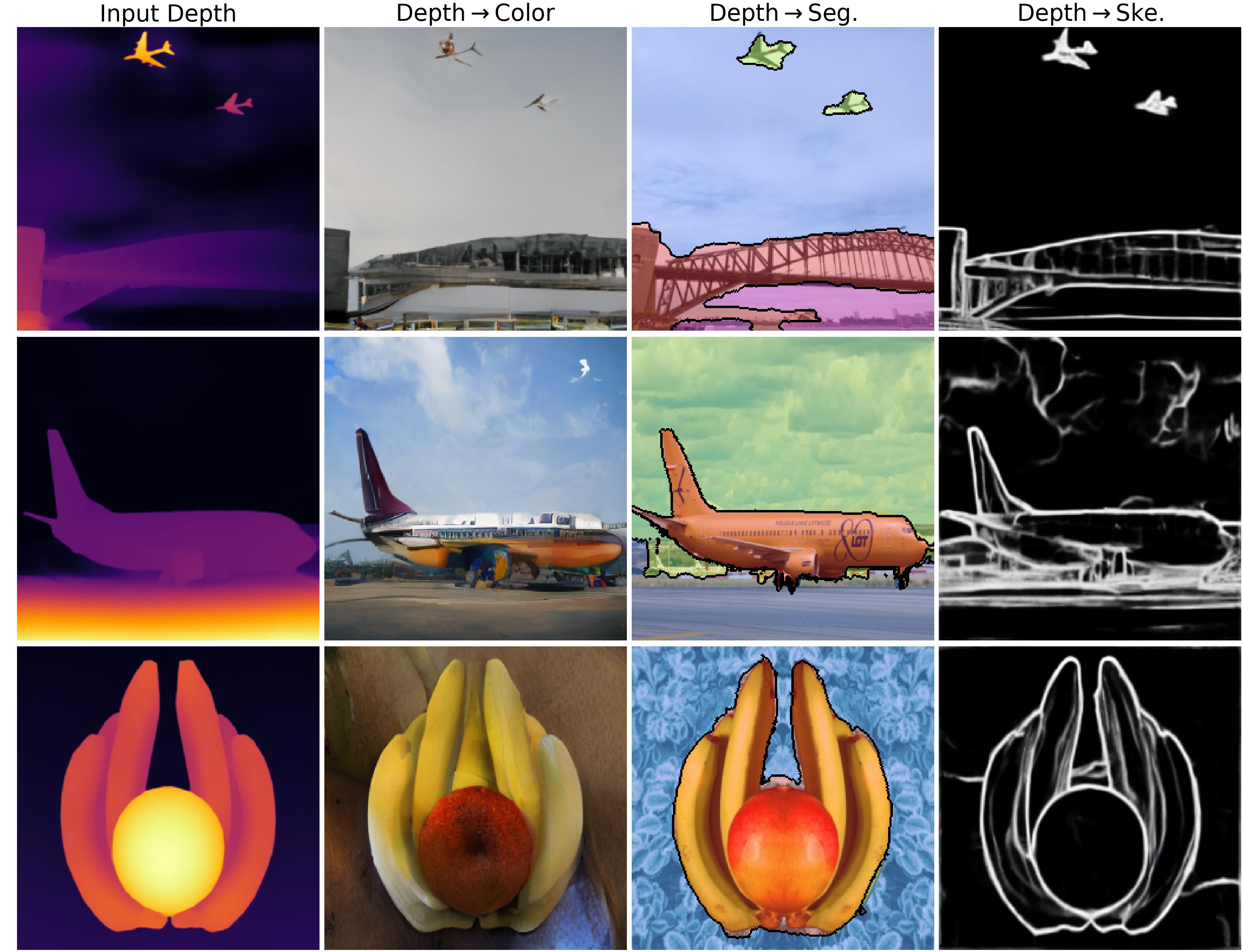}
\par\end{centering}
}
\par\end{centering}
\centering{}\caption{Results of finetuned dDR on COCO-UMDT-Star (a) and COCO-UMDT-Chain
(b) for translation tasks from Depth domain to Color, Segment, and
Sketch domains.\label{fig:dDR_coco_depth}}
\end{figure}

\begin{figure}
\begin{centering}
\subfloat[Results on COCO-UMDT-Star. The classes for each segmentation map (from
top to bottom) are: (\textcolor{green}{bed}\textcolor{black}{, }\textcolor{magenta}{floor}\textcolor{black}{,}\textcolor{red}{{}
}\textcolor{purple}{window}, \textcolor{cyan}{light}, \textcolor{orange}{wall}),
(\textcolor{blue}{pizza}, \textcolor{cyan}{undefined}, \textcolor{green}{table},
\textcolor{red}{fork}), (\textcolor{green}{grass}, \textcolor{red}{giraffe},
\textcolor{blue}{bush}, \textcolor{magenta}{sky}).]{\begin{centering}
\includegraphics[width=0.95\textwidth]{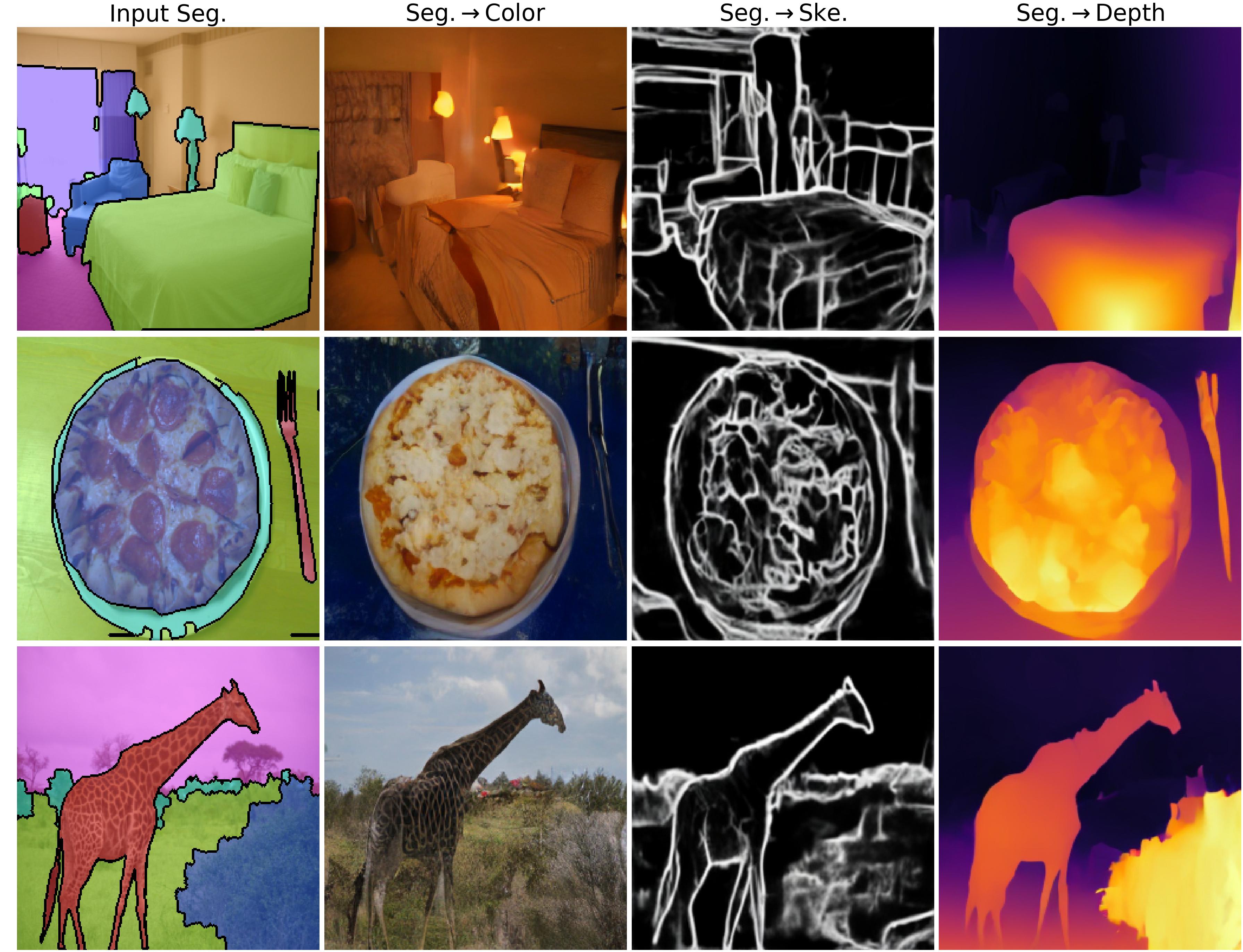}
\par\end{centering}
}
\par\end{centering}
\begin{centering}
\subfloat[Results on COCO-UMDT-Chain. The classes for each segmentation map
(from top to bottom) are: (\textcolor{green}{bed}\textcolor{black}{,
}\textcolor{magenta}{floor}\textcolor{black}{,}\textcolor{red}{{} }\textcolor{purple}{window},
\textcolor{cyan}{light}, \textcolor{orange}{wall}), (\textcolor{blue}{pizza},
\textcolor{cyan}{undefined}, \textcolor{green}{table}, \textcolor{red}{fork}),
(\textcolor{green}{grass}, \textcolor{red}{giraffe}, \textcolor{blue}{bush},
\textcolor{magenta}{sky}).]{\begin{centering}
\includegraphics[width=0.95\textwidth]{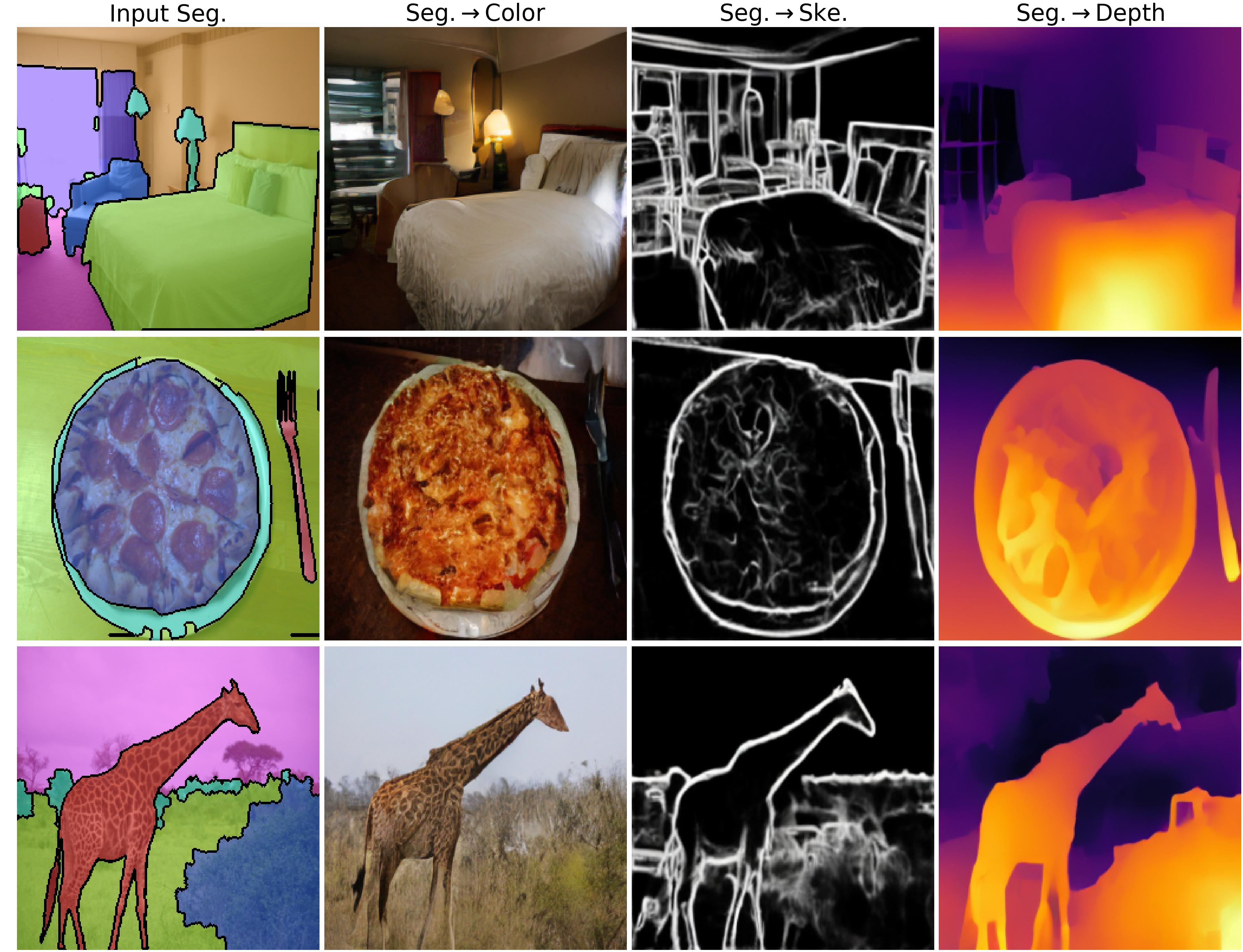}
\par\end{centering}
}
\par\end{centering}
\centering{}\caption{Results of finetuned dDR on COCO-UMDT-Star (a) and COCO-UMDT-Chain
(b) for translation tasks from Segment domain to Color, Sketch, and
Depth domains.\label{fig:dDR_coco_seg}}
\end{figure}

We visualize dDR's results on COCO-UMDT-Star and COCO-UMDT-Chain to
illustrate its ability to handle UMDT. Specifically, Figs.~\ref{fig:dDR_coco_color},
\ref{fig:dDR_coco_ske}, \ref{fig:dDR_coco_depth}, and \ref{fig:dDR_coco_seg}
show cross-domain samples conditioned on color images, sketches, depth
maps, and segmentation maps, respectively. Because segmentation maps
are not ideally suited to a continuous diffusion process in latent
space and predicted class labels are hard to decode perfectly, we
apply DBSCAN \cite{ester1996a} as a post-processing step that merges
noisy pixels into the nearest region with the smallest label difference.

